\documentclass[conference,a4paper]{IEEEtran}
\IEEEoverridecommandlockouts

\usepackage{siunitx}
\usepackage{cite}
\usepackage{amsmath,amssymb,amsfonts}
\usepackage{graphicx}
\usepackage{textcomp}
\usepackage{xcolor}
\usepackage{subcaption}
\usepackage{algorithm}
\usepackage{algpseudocode}
\usepackage[acronym]{glossaries}
\usepackage{hyperref}
\usepackage{ifthen}
\usepackage{textgreek}
\usepackage[acronym]{glossaries}
\usepackage{amsmath}
\usepackage{amsfonts}
\usepackage{ifthen}
\usepackage{gensymb}
\def\BibTeX{{\rm B\kern-.05em{\sc i\kern-.025em b}\kern-.08em
    T\kern-.1667em\lower.7ex\hbox{E}\kern-.125emX}}
\newacronym{OMPOSO}{OMPOSO}{optimized multiobjective particle swarm optimization}
\newacronym{AOA}{AOA}{angle of arrival}
\newacronym{FMCW}{FMCW}{frequency modulated continuous wave}
\newacronym{PSO}{PSO}{particle swarm optimization}
\newacronym{AMCL}{AMCL}{adaptive Monte Carlo localization}
\newacronym[plural = AMRs]{AMRs}{AMR}{autonomous mobile robot}
\newacronym{RCS}{RCS}{radar cross section}
\newacronym{LOS}{LOS}{line of sight}
\newacronym{CRLB}{CRLB}{Cramer-Rao lower bound}
\newacronym{GDOP}{GDOP}{geometric dilution of precision}
\newacronym[plural = LRPs]{LRPs}{LRP}{local reference point}
\newacronym{FIM}{FIM}{Fisher information matrix}
\newacronym{MO}{MO}{multi-objective}
\newacronym{MLAT}{MLAT}{multilateration}
\newacronym{GPU}{GPU}{graphics processing unit}
\newacronym{GA}{GA}{genetic algorithm}
\newacronym{RF}{RF}{radio frequency}
\newacronym{UWB}{UWB}{ultra wideband}
\newacronym{TOA}{TOA}{time of arrival}
\newacronym{TDOA}{TDOA}{time difference of arrival}
\newacronym{IPS}{IPS}{indoor positioning system}
\newacronym{SLAM}{SLAM}{simultaneous localization and mapping}
\newacronym{RMSE}{RMSE}{root-mean-square error}
\glsaddall

\makeatletter
\let\old@ps@headings\ps@headings
\let\old@ps@IEEEtitlepagestyle\ps@IEEEtitlepagestyle
\def\confheader#1{%
	\def\ps@headings{%
		\old@ps@headings%
		\def\@oddhead{\strut\hfill#1\hfill\strut}%
		\def\@evenhead{\strut\hfill#1\hfill\strut}%
	}%
	\def\ps@IEEEtitlepagestyle{%
		\old@ps@IEEEtitlepagestyle%
		\def\@oddhead{\strut\hfill#1\hfill\strut}%
		\def\@evenhead{\strut\hfill#1\hfill\strut}%
	}%
	\ps@headings%
}
\makeatother

\begin{document}

\title{Indoor Positioning based on Active Radar Sensing and Passive Reflectors: Reflector Placement Optimization}
\author{\IEEEauthorblockN{Sven Hinderer, Pascal Schlachter, Zhibin Yu, Xiaofeng Wu, Bin Yang}
\author{\IEEEauthorblockN{Sven Hinderer*, Pascal Schlachter*, Zhibin Yu$\dag$, Xiaofeng Wu$\dag$, Bin Yang*}
\IEEEauthorblockA{\textit{*Institute of Signal Processing and System Theory, University of Stuttgart, Stuttgart, Germany} \\
\textit{$\dag$Munich Research Center, Huawei Technologies Duesseldorf GmbH, Munich, Germany}\\
firstname.lastname@iss.uni-stuttgart.de, zhibinyu@huawei.com, xiaofeng.wu1@huawei.com}
}}
\maketitle

\noindent\textbf{Note:} This is the accepted version of the paper. 
The final version is published in the \emph{Proceedings of the IEEE 2023 13th International Conference on Indoor Positioning and Indoor Navigation (IPIN)}. 
DOI: \href{https://doi.org/10.1109/IPIN57070.2023.10332502}{10.1109/IPIN57070.2023.10332502}

\vspace{1em}

\renewcommand{\thefootnote}{}
\footnotetext{
	\textcopyright~2023 Personal use of this material is permitted.  Permission from IEEE must be obtained for all other uses, in any current or future media, including reprinting/republishing this material for advertising or promotional purposes, creating new collective works, for resale or redistribution to servers or lists, or reuse of any copyrighted component of this work in other works
}
\renewcommand{\thefootnote}{\arabic{footnote}}

\begin{abstract}
We extend our work on a novel~\gls{IPS} for~\glspl{AMRs} based on radar sensing of local, passive radar reflectors. Through the combination of simple reflectors and a single-channel~\gls{FMCW} radar, high positioning accuracy at low system cost can be achieved. Further, a~\gls{MO}~\gls{PSO} algorithm is presented that optimizes the 2D placement of radar reflectors in complex room settings.
\end{abstract}


\section{Introduction}
Currently, most high accuracy~\gls{RF} based indoor localization systems apply the~\gls{UWB} approach. While~\gls{UWB} is an effective technology, it suffers from high cost~\cite{uwb}.

\gls{FMCW} radars provide a cheap alternative, as digitization and processing of the received signal takes place after down-conversion to baseband. In combination with reflective tags that modulate the incident wave, which makes the tags identifiable, high positioning accuracy is achievable~\cite{millimetro, fmcwindoor}. However, this requires active reflectors and, for positioning with a single tag, beamforming and therefore a multi-channel system. 

We present a single-channel~\gls{FMCW}~\gls{IPS} that works with passive reflectors, offering a favorable accuracy to system cost ratio. The system has a fingerprinting and a~\gls{MLAT} mode. We build on our previous work~\cite{pascal} including a positioning and tracking solution based on~\gls{AMCL}~\cite{prob_robotics}, which proved the fingerprinting concept in a small room under the assumption that all reflectors can be detected by the~\gls{AMRs}. Our previous work used heuristic rules for the reflector placement to improve the global localization performance using fingerprinting for positioning.

This work extends the concept to arbitrary, larger rooms with limited visibility of reflectors. We focus on the development of a novel, advanced PSO algorithm to optimize the reflector placement for both positioning modes. The gain in performance with optimized LRP placements is demonstrated by evaluating the positioning accuracy using the~\gls{AMCL} algorithm.

\section{System assumptions}
Our conceptual system consists of a~\gls{FMCW} radar with limited range resolution $r_{res}$ placed on top of an~\gls{AMRs}. The radar senses passive reflectors, denoted also~\glspl{LRPs}, mounted at known positions in the room above the radar, as depicted in Fig.~\ref{fig:cone}.

Positioning can be performed in two modes. The less accurate fingerprinting mode allows for global localization. If each position in the room can be assigned a unique localization fingerprint that is inferred from the detected LRPs, the~\gls{AMRs} position can be easily estimated by matching the measured fingerprint with a pre-computed fingerprint database of the room. With simple, passive reflectors, we assume that we can at most differentiate two different LRP types by e.g. their~\gls{RCS}. This restriction leads to~\gls{AMRs} positions that share the same fingerprint, thus introducing fingerprint ambiguity and impairing the localization accuracy~\cite{pascal}. The integration of angular or polarimetric information of LRPs in the fingerprint will alleviate this problem, but leads to a higher radar system complexity. This extension is not investigated in this study. 

Second, we introduce an~\gls{MLAT} mode.~\gls{MLAT} requires a distinct identifier for each~\gls{LRPs}. Once we approximately know the AMR position through fingerprinting, we can use this information in conjunction with the known LRP placement to assign LRP identifiers to the estimated AMR-LRP distances. The known LRPs and their distances to the AMR are then used to localize the AMR via~\gls{MLAT}. This can be realized by tracking the LRPs in a~\gls{SLAM} like fashion. The difference to conventional~\gls{SLAM} is that we already know the~\gls{LRPs} map, which simplifies the problem and allows for absolute instead of  relative position estimates.

\section{Problem formulation}
\label{sec:problem}
\subsection{Notations}
To optimize the LPR placement, we search for a trade-off between both positioning modes, which have different LRP placement goals. Before formulating the respective optimization objectives, we define the room.

The room is modeled as a 2D area $P_{room}\subset\mathbb{R}^2$ with a polygon boundary. A radar is placed at a fixed height $z_r$ on the~\gls{AMRs}. We build a planar 2D grid with quadratic grid elements of given size in the $x$\nobreakdash-$y$~plane and define the set of their 3D center locations as $\mathcal{I}_{c}$. The set of 3D grid element center positions that are in the room is $\mathcal{I}_{grid}=\{{\underline{p}_r}| [x_r,y_r]^T \in P_{room}\wedge\underline{p}_r\in \mathcal{I}_{c}\}$ with $\underline{p}_r=[x_r, y_r, z_r]^T\in\mathbb{R}^3$, where $\underline{x}^T$ denotes the transpose of a vector $\underline{x}$. This represents the finite set of evaluated AMR positions with a fixed radar height $z_r$. A~\gls{LRPs} $l$ has a 3D position $\underline{q}_l=[x_l, y_l, z_l]^T\in\mathbb{R}^3$. The set of positions of all $M$ LRPs is $\mathcal{I}_{LRPs}=\{\,{\underline{q}_l}, 1\leq l\leq M \,\}$. Currently, the height $z_l$ at which LRPs are placed is fixed. Consequently, only the undiscretized $x_l$ and $y_l$ coordinates of~\glspl{LRPs} are to be optimized.
\begin{figure}[h!]
	\centering
	\includegraphics[width=0.6\linewidth]{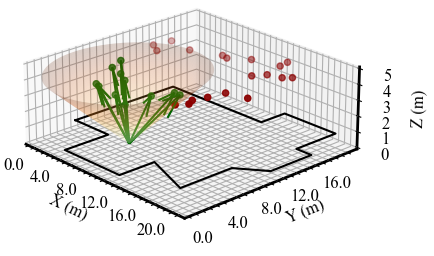}
	\caption{Our conceptual system. A radar (at the cone vertex) senses LRPs (dots, green dots are detected) above it and uses them for localization.}
	\label{fig:cone}
\end{figure}
\subsection{Fingerprinting objective}
\begin{figure}[htp]
	\centering
	\includegraphics[width=0.6\linewidth]{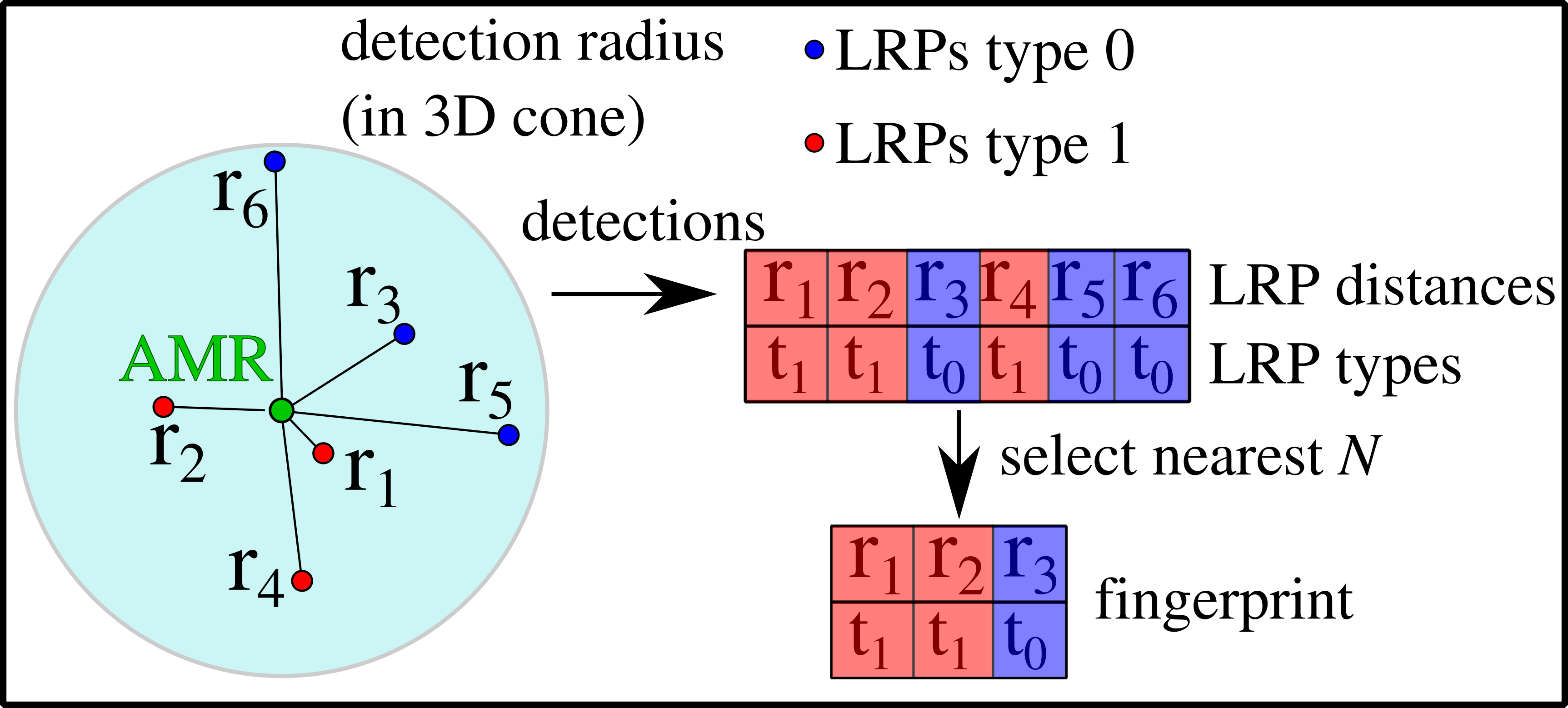}
	\caption{Top-down view of the fingerprint calculation. The nearest $N$ (here $N=3$) detected LRPs and corresponding LRP types give the fingerprint for the AMR position.}
	\label{fig:fingerprint}
\end{figure}

In this study, we use a fixed number of $N<M$ LRPs in a fingerprint for simplicity. This assumption will be relaxed in future work. To calculate the fingerprint of a grid element, we compute the distances from the grid center location $\underline{p}_r$ to all visible~\glspl{LRPs}. As our radar has a limited range resolution $r_{res}$, these distances are rounded with the resolution $r_{res}$. We then select the nearest $N$ detected~\glspl{LRPs}. The set of their $N$ rounded distances and corresponding $N$ LRP types is the fingerprint $\mathcal{F}_r$ for the considered grid element center position $\underline{p}_r$. The mapping from $\underline{p}_r$ to its fingerprint is described by the function
\begin{align}
	\mathcal{F}_r = g(\underline{p}_r),
\end{align}
where the dependence of $g()$ on the LRP placement is dropped for simplicity. An example of a fingerprint calculation is shown in Fig.~\ref{fig:fingerprint}.

We optimize the LRP placement for AMR positioning in the fingerprinting mode by minimizing the number of ambiguous grid elements. A grid element is defined ambiguous, if there exists at least another grid element with the same fingerprint. Let $\mathcal{I}_{amb}$ be the set of all ambiguous grid elements, then our fingerprinting objective $f_1$ becomes 
\begin{align}
	f_1 = |\mathcal{I}_{amb}|.
\end{align}
Minimizing $f_1$ reduces the fingerprinting ambiguity and enhances both local and global~\gls{AMRs} positioning accuracy.

\subsection{MLAT objective}
A common metric to evaluate the performance of~\gls{MLAT} systems is the~\gls{GDOP}. It is known from~\cite{gdop3drange} that the~\gls{GDOP} for a position $\underline{p}_r$ in a~\gls{MLAT} positioning system based on absolute range measurements under the assumption of uncorrelated Gaussian range estimation errors with zero mean and variance $\sigma_r^2$ (here assumed $\sigma_r=r_{res}$) is given by
\begin{align}
	\mathrm{GDOP}(\underline{p}_r) = \mathrm{tr}\left(\left(\mathbf{H}_r^T\mathbf{H_r}\right)^{-1}\right)\sigma_r^2.
\end{align}
tr($\mathbf{X}$) denotes the trace of a matrix $\mathbf{X}$, $\mathbf{X}^{\textrm{-1}}$ its inverse and  $\mathbf{J}_r=\mathbf{H}_r^T\mathbf{H}_r$ is the~\gls{FIM}. The matrix $\mathbf{H}_r$ contains the geometric information of our LRP placement and is calculated as

\begin{align}
	\mathbf{H}_r = [\underline{h}_{r,1},\,\dots\,,\,\underline{h}_{r,K}]^T \quad\mathrm{with}\quad
	\underline{h}_{r,s} = \frac{\underline{p}_r-\underline{q}_s}{\lVert\underline{p}_r-\underline{q}_s \rVert}.
\end{align}
It is assumed that only a subset $\mathcal{I}_{K}\subset\mathcal{I}_{LRPs}$ of $|\mathcal{I}_{K}|=K$ LRPs at the positions $\underline{q}_1,\,\dots\,,\,\underline{q}_K$ are visible from $\underline{p}_r$. We define our~\gls{MLAT} objective $f_2$ as the sum of the~\gls{GDOP} over all grid elements in the room
\begin{align}
	f_2 = \sum_{\underline{p}_r\in\mathcal{I}_{grid}}{\mathrm{GDOP}(\underline{p}_r)}.
\end{align}
Minimizing $f_2$ improves the positioning accuracy of the~\gls{MLAT} mode for~\gls{AMRs} positioning, which gives higher local positioning accuracy than fingerprinting, but cannot be used for global localization.
\subsection{Overall optimization objectives}
We formulate the optimization problem of the LRP placement as
\begin{align}
	&\underline{\hat{\theta}}=\min_{\underline{\theta}\in\mathbf{\Theta}}\left\{f_1, f_2\right\} \\
	\textrm{s.t.}\, & M\leq M_{max}\label{eq:max_LRPs}\\ 
					& K \geq 4\label{eq:coverage}\\ 
					&  \lVert\underline{q}_i-\underline{q}_j\rVert \geq\SI{0.5}{m}\,\forall i\neq j\label{eq:min_distance}\\ 
					&\underline{q}_l\in P_{b}\label{eq:roomboundary}\,\forall l.
\end{align}
$\Theta$ is the set of all feasible 2D LRP placements and $\underline{\theta}=[x_1, y_1,\,\dots\,,\,x_M, y_M]^T$ holds the $x$ and $y$ coordinates of all $M$~\glspl{LRPs}. The first boundary condition bounds the maximum number of~\glspl{LRPs} by $M_{max}$. The second condition ensures  a minimum number of 4 visible~\glspl{LRPs} in all evaluated grid elements. The third condition denotes a minimum distance of \SI{50}{cm} between all~\glspl{LRPs}, which facilitates the LRP installation and is required to reliably differentiate between them in the detection process. With the last condition,~\glspl{LRPs} have to be in the room area $P_{b}\subset P_{room}$ that is built from $P_{room}$ with a \SI{50}{cm} distance to the room boundary. This keeps all LRPs inside the room with a minimum distance of 50 cm to the nearest wall.
To solve the above nonlinear, nondifferentiable,~\gls{MO} optimization problem, we propose an advanced~\gls{PSO} algorithm in the following section.

\section{Optimum LRP placement with PSO}
\label{sec:pso}
\subsection{Particles and swarm}
\gls{PSO} is a metaheuristic optimization algorithm based on the swarm behavior of multiple particles~\cite{pso}. A particle $j$ has a position $\underline{\theta}_j$\footnote{The position vector $\underline{\theta}_j$ has nothing to do with the AMR or LRP positions. It is a term in swarm optimization and contains all parameters to be optimized.}, denoting a candidate solution, i.e. a 2D LRP placement in our problem. It also has a \emph{pbest} solution, which is the best position vector that this particle has ever found during the iterative optimization. The whole set of particles, referred to as swarm, shares a \emph{gbest} solution. Similarly, \emph{gbest} holds the best parameters that all particles have found. During each~\gls{PSO} iteration $i$, a particle $j$ is updated by first calculating its velocity vector $\underline{v}_{j}(i)=[v_{x_1}, v_{y_1},\,\dots\,,\,v_{x_M}, v_{y_M}]^T$ with~\cite{omopso}

\begin{equation}
\begin{aligned}	
	\underline{v}_{j}(i) =  W\underline{v}_{j}(i-1)\:&+C_{1}r_{1}(\underline{\theta}_{pbest,j}-\underline{\theta}_{j}(i-1))\\\:&+ C_{2}r_2(\underline{\theta}_{gbest}-\underline{\theta}_{j}(i-1))
	\label{eq:velocity}.
\end{aligned}
\end{equation}
The first term is the velocity from the previous iteration, scaled with the inertia weight $W$. The second term pushes the particle $j$ towards its~\emph{pbest} solution and has a random factor $r_1$ and cognitive factor $C_1$. It corresponds to the cognitive behavior of a particle that uses its best solution as a leader LRP placement to guide the search. The third term moves the particle $j$ towards the shared~\emph{gbest} solution, which has another random factor $r_2$ and a social factor $C_2$. After the new velocity vector is computed, the particle position is updated using
\begin{align}
	\underline{\theta}_{j}(i) = \underline{\theta}_{j}(i-1) + \underline{v}_{j}(i).
	\label{eq:position}
\end{align}

We only optimize the 2D coordinates of LRPs in the placement but save the full placement (3D + types) in a particle, as it is required for AMR positioning. We also largely extend this simple PSO variant as below, as it cannot effectively deal with the complex problem at hand.

Currently, the LRP placement is designed such that different particles share the same number of LRPs for each type $t\in\left\{0,1\right\}$. We distribute LRPs of each type equally in a LRP placement. If we use two LRP types and a particle has an odd number of LRPs, we use one more LRP of type 0 than type 1.

\subsection{Multiobjective PSO}
To optimize the LRP placement for both the fingerprinting and~\gls{MLAT} mode, we leverage the core idea from~\gls{OMPOSO}~\cite{omopso}, which shows favorable performance compared with other~\gls{MO} PSO variants or the widely used~\gls{MO}~\gls{GA} NSGA-II~\cite{nsga2, nsga2vsomopso} on various benchmark problems. OMOPSO optimizes a set of $m$ objectives $\left\{f_i(\underline{\theta}),\,1 \leq i\leq m \right\}$ using Pareto dominance. A solution $\underline{\theta}_1$ is said to dominate another solution $\underline{\theta}_2$ if
\begin{align}
	&f_i(\underline{\theta}_1)\le f_i(\underline{\theta}_2), \forall i=1,\,\dots\,,\,m\,\, \mathrm{and}\\
	&f_i(\underline{\theta}_1)<f_i(\underline{\theta}_2), \exists i=1,\,\dots\,,\,m.
\end{align}
It has to be at least equally good in all objectives and strictly better in at least one objective to dominate. If $\underline{\theta}_1$ does not dominate $\underline{\theta}_2$, $\underline{\theta}_2$ is also said to be non-dominated. The goal is then to approximate the so called Pareto optimal set, the unknown set of Pareto optimal solutions. 
\subsection{Variable number of dimensions}
OMOPSO cannot deal with a variable number of dimensions, i.e. $\dim(\underline{\theta})$ is constant and cannot be minimized as a third objective. This means the number of LRPs in the LRP placements is fixed, as their 2D coordinates are contained in the solution $\underline{\theta}_j$. In order to reduce the required number of~\glspl{LRPs}, we extend the basic~\gls{OMPOSO} algorithm by mutation operations that add or remove~\glspl{LRPs} from a particle. We call them \emph{upmutation} and \emph{downmutation}. Mutating the dimension was proposed in~\cite{vndpso} for standard~\gls{PSO}.

\emph{Upmutation}: If we don't cause a violation of Eq.~\ref{eq:max_LRPs} after \emph{upmutation}, a new LRP is placed at a random position inside $P_{b}$ and appended to $\underline{\theta}_j$. A corresponding random 2D velocity vector is created and appended to $\underline{v}_j$.

\begin{figure}[h!]
	\centering
	\includegraphics[width=0.9\linewidth]{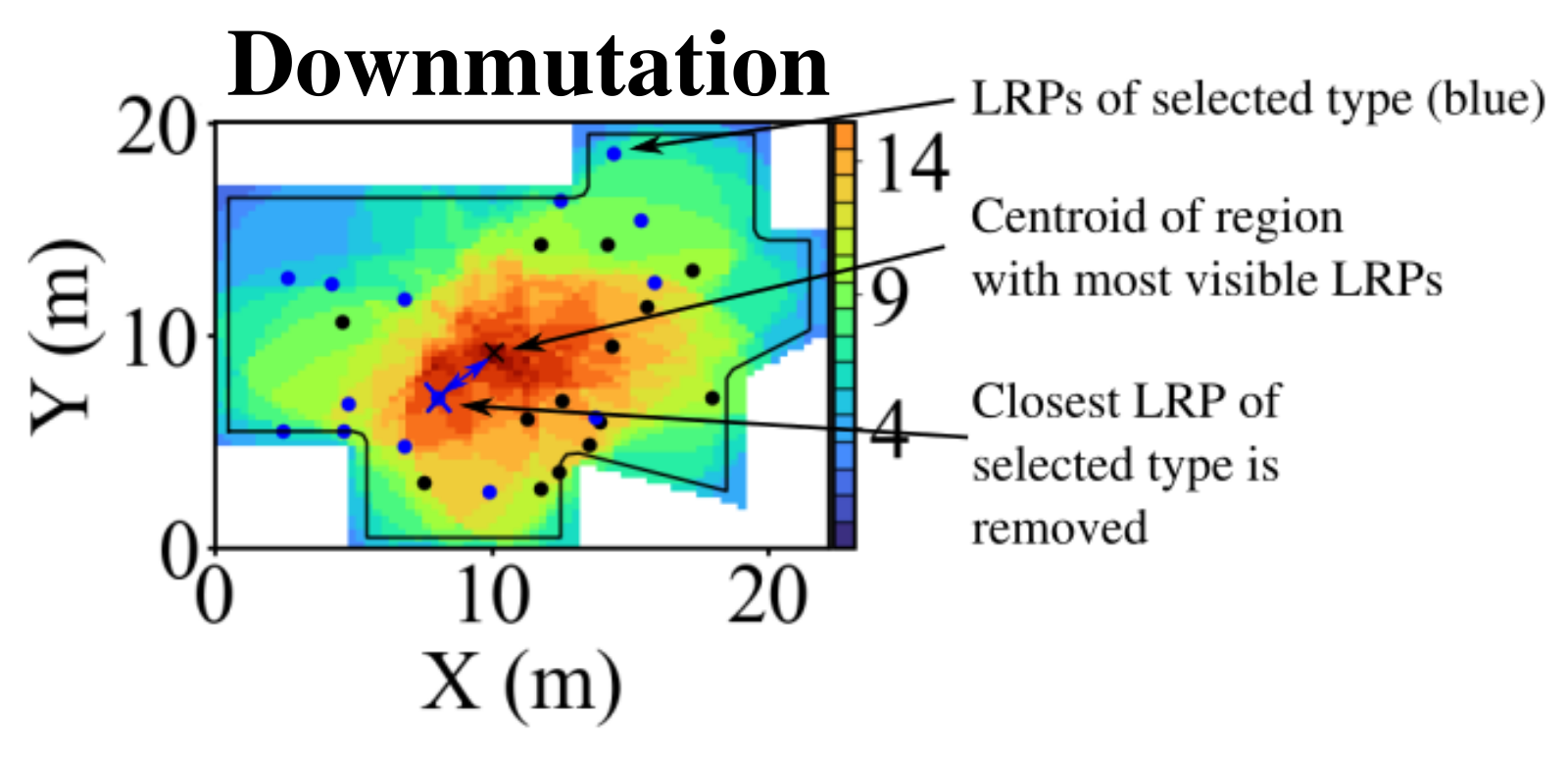}[t]
	\caption{By removing the LRP of selected type where most LRPs are visible, we avoid coverage violations after \emph{downmutation}. The color indicates the number of visible LRPs.}
	\label{fig:downmutation}
\end{figure}

\emph{Downmutation}: Random LRP deletion would create many coverage violations. Instead, we search for the cluster of grid elements in $\mathcal{I}_{grid}$ where most~\glspl{LRPs} of all types are visible and calculate its centroid. Then, we select the LRP type that shall be removed. This can be inferred from the described equal distribution of LRPs of each type. The LRP of this type that is closest to the centroid  and its velocity are then removed from the particle. The procedure is shown in Fig.~\ref{fig:downmutation}.
\subsection{Boundary conditions}
The polygon boundary condition Eq.~\ref{eq:roomboundary} for LRPs can be handled in the standard~\gls{PSO} fashion by projecting LRPs outside of $P_{b}$ onto their nearest point on the boundary of $P_{b}$. The coverage in Eq.~\ref{eq:coverage} and the distance between~\glspl{LRPs} in Eq.~\ref{eq:min_distance} are harder to enforce. One option would be to add penalty terms for boundary condition violations to the objective functions. Instead, we propose two physically inspired models that can project an unfeasible LRP placement back to a feasible one. Similar ideas have been used in~\cite{vforce}, where attractive and repulsive virtual forces are applied to ranging sensors as a deployment strategy to maximize the region they cover. Integrating these physically motivated models in the optimization routine notably improves the quality of solutions. They are also used for the initialization of particles, as the standard PSO method of random LRP placement cannot produce feasible solutions in many cases.

\begin{figure}[h!]
	\centering
	\begin{subfigure}[b]{.225\textwidth}\
		\includegraphics[width=\textwidth]
		{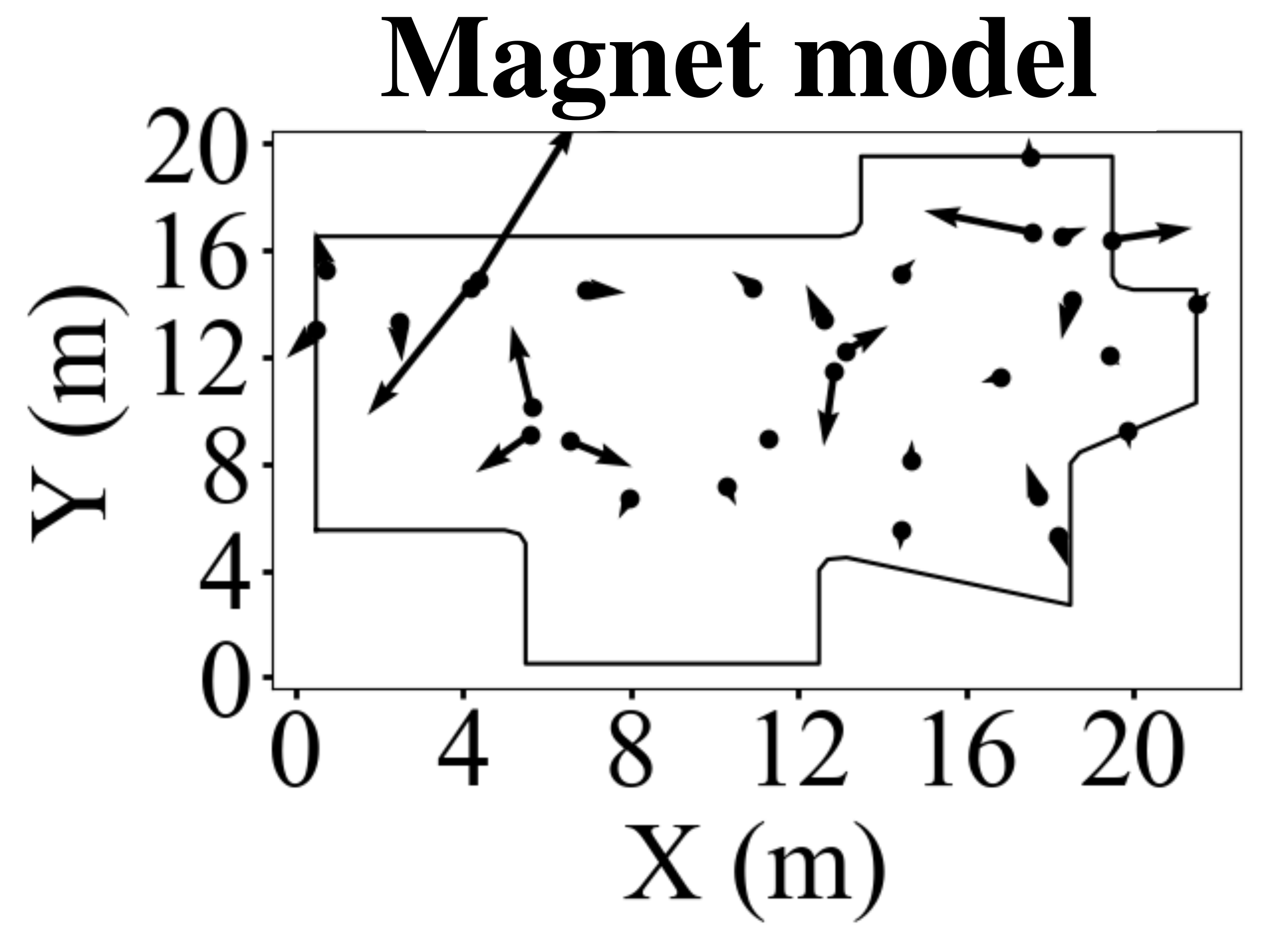}
		\caption{LRPs act like magnets of the same polarity.}
		\label{fig:mag}
	\end{subfigure}
	\hfill
	\begin{subfigure}[b]{.255\textwidth}
		\includegraphics[width=\textwidth]{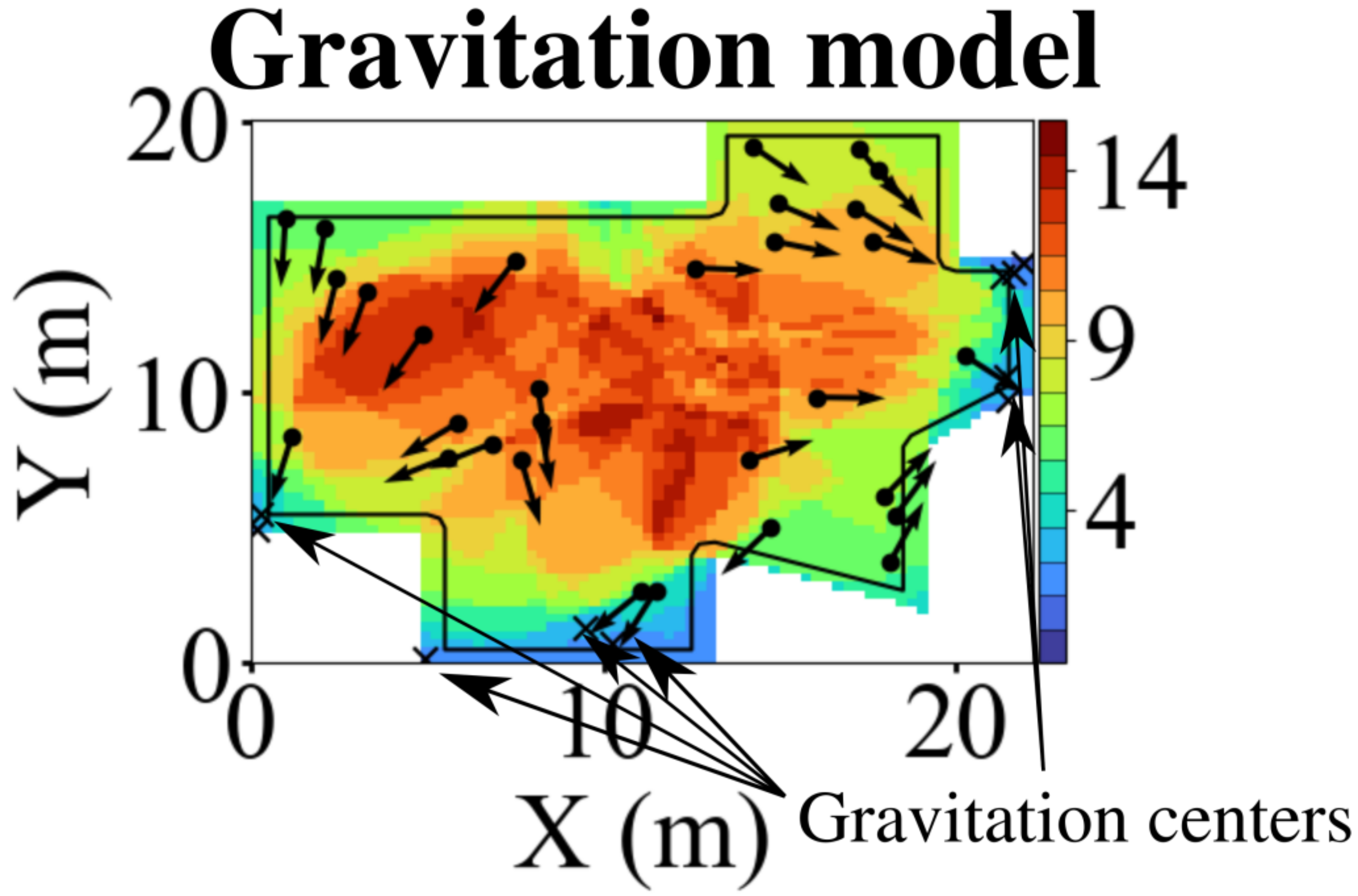}
		\caption{LRPs gravitate towards regions with coverage violations.}
		\label{fig:grav}
	\end{subfigure}
	\caption{Our physically inspired models to handle and fix boundary condition violations.}
	\label{fig:b_models}
\end{figure}

To assert \SI{50}{cm} distance between~\glspl{LRPs}, we apply a model that treats LRPs like magnets of the same polarity. If they come too close, they will push each other away in opposing directions, as shown in Fig.~\ref{fig:mag}. 

For the coverage problem, we implement a model inspired by gravitation. We calculate the centroids of all regions in $\mathcal{I}_{grid}$ where the coverage is violated. These positions then act as centers of gravitation and~\glspl{LRPs} are pulled towards their closest gravitation center. A plot depicting the gravitation model is given in Fig.~\ref{fig:grav}.

\begin{figure}[h!]
	\centering	
	\begin{subfigure}[t]{.155\textwidth}
		\centering
		\includegraphics[width=\textwidth]{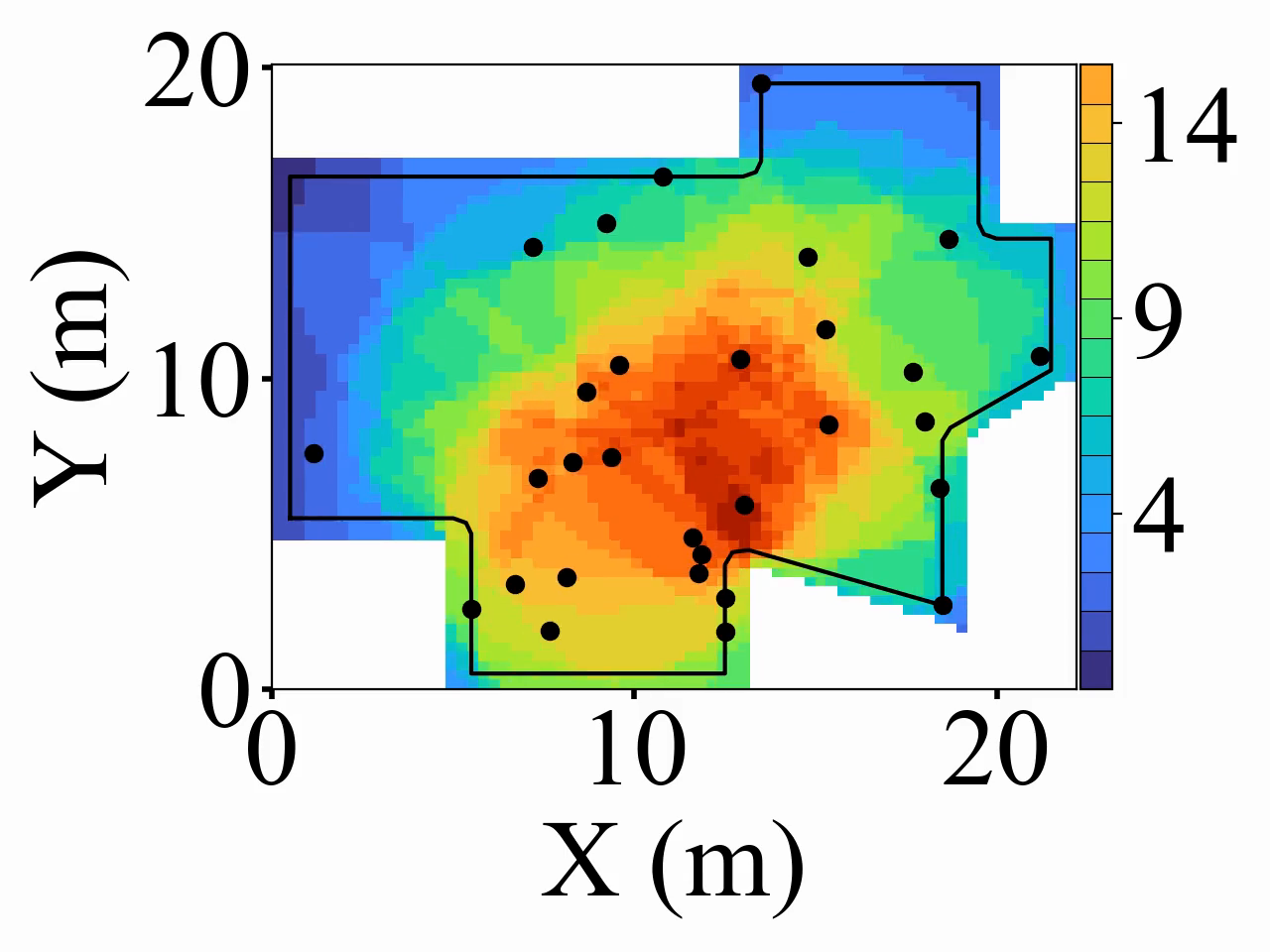}
		\caption{Initial iteration 1}
		\label{fig:it1}
	\end{subfigure}
	\begin{subfigure}[t]{.155\textwidth}
		\centering
		\includegraphics[width=\textwidth]{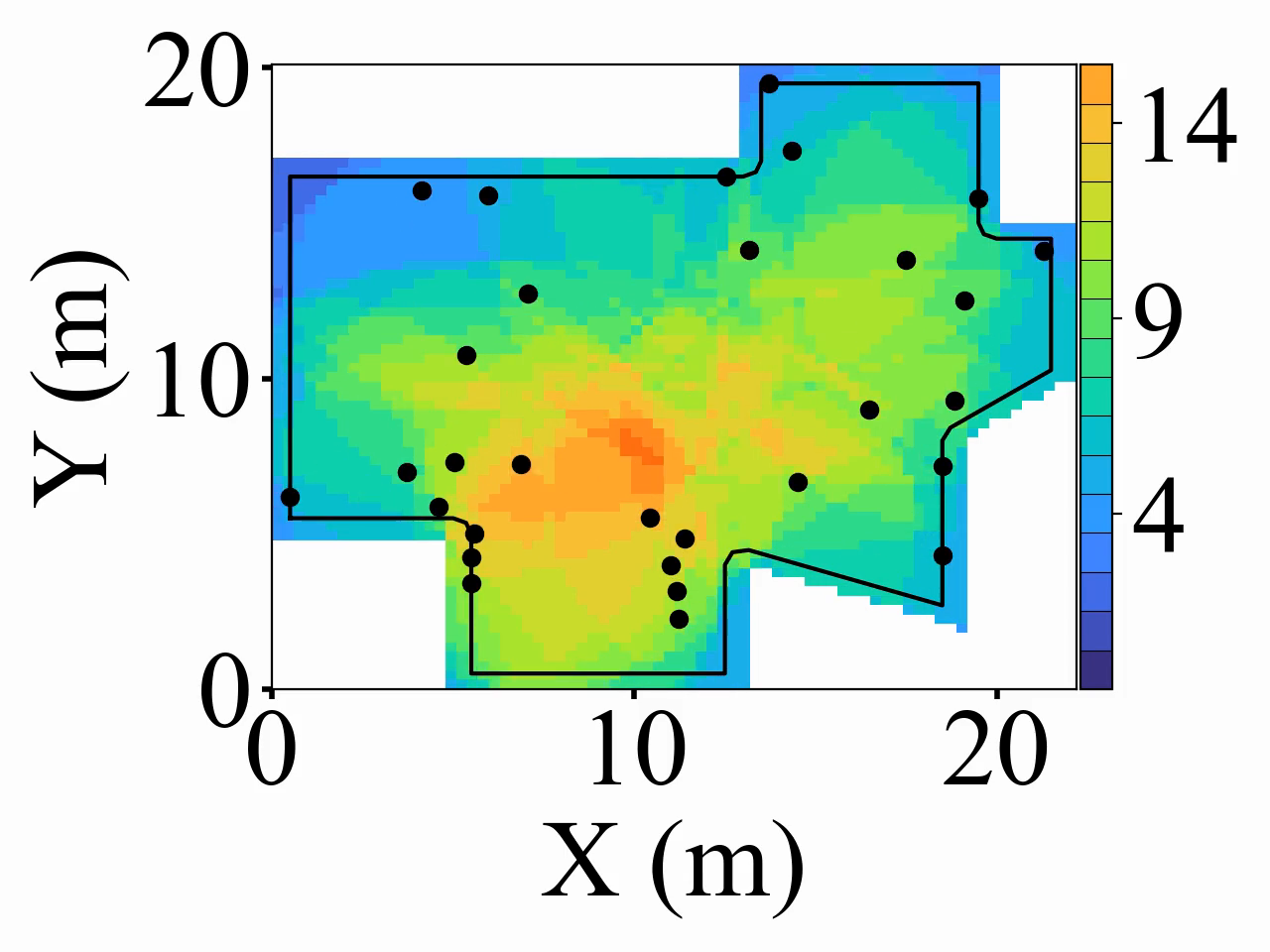}
		\caption{Iteration 12}
		\label{fig:it12}
	\end{subfigure}
	\begin{subfigure}[t]{.155\textwidth}
		\centering
		\includegraphics[width=\textwidth]{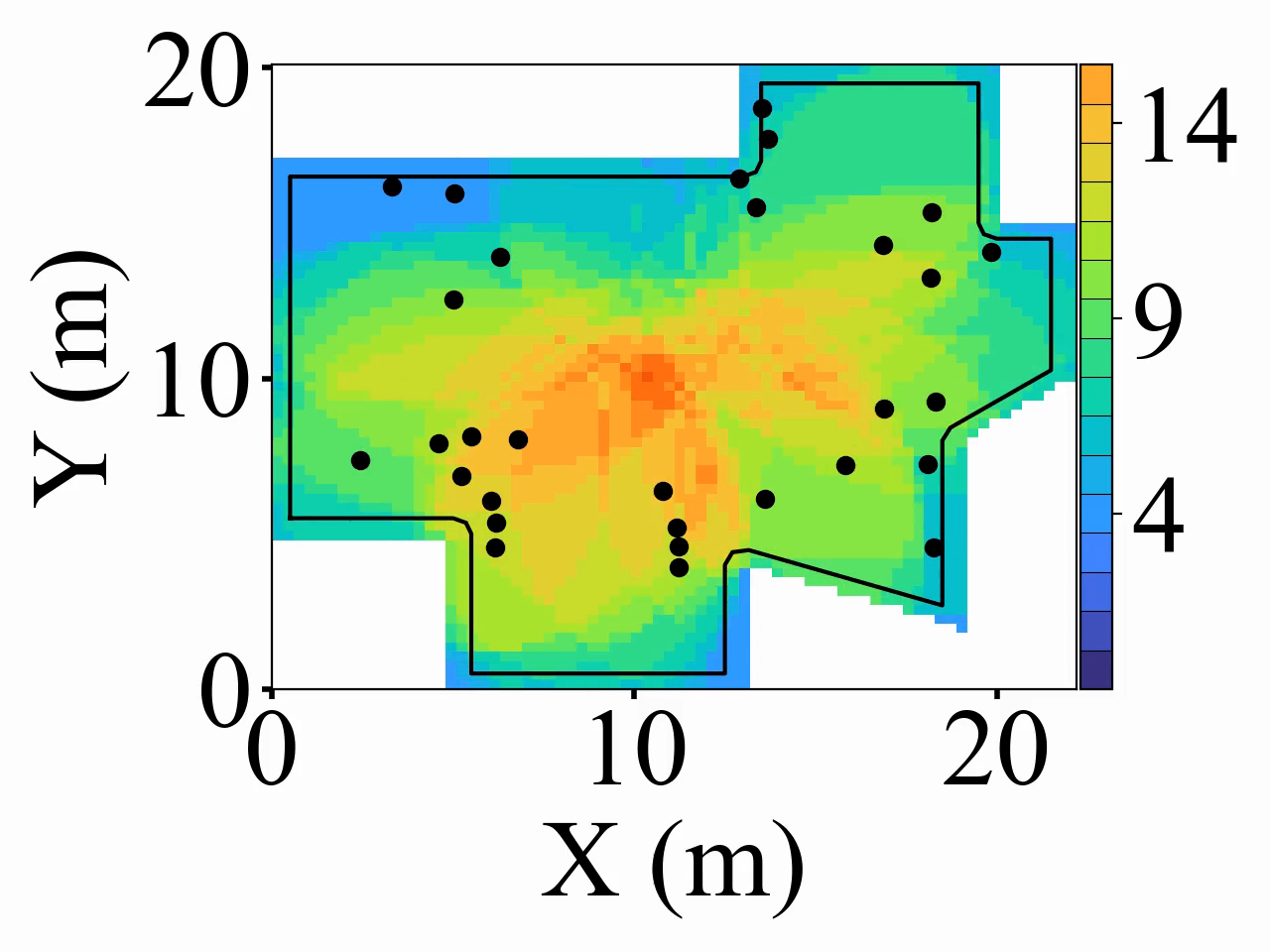}
		\caption{Iteration 26 (feasible)}
		\label{fig:it26}
	\end{subfigure}
	\caption{Iterative boundary condition subroutine. LRPs are drawn towards regions with coverage violations and repulse each other. After a few iterations, a feasible solution is found. This also works if the initial solution is far away from being feasible.}
	\label{fig:bound}
\end{figure}

When iteratively applying the \emph{magnet model}, \emph{gravitation model} and the projection of LRPs outside of $P_b$ onto its boundary, a feasible solution can be quickly reached in most cases, as shown in Fig.~\ref{fig:bound}.

\subsection{Permutation invariant velocity update}
\begin{figure}[h!]
	\centering
	\begin{subfigure}[htp]{.23\textwidth}\
		\includegraphics[width=\textwidth]
		{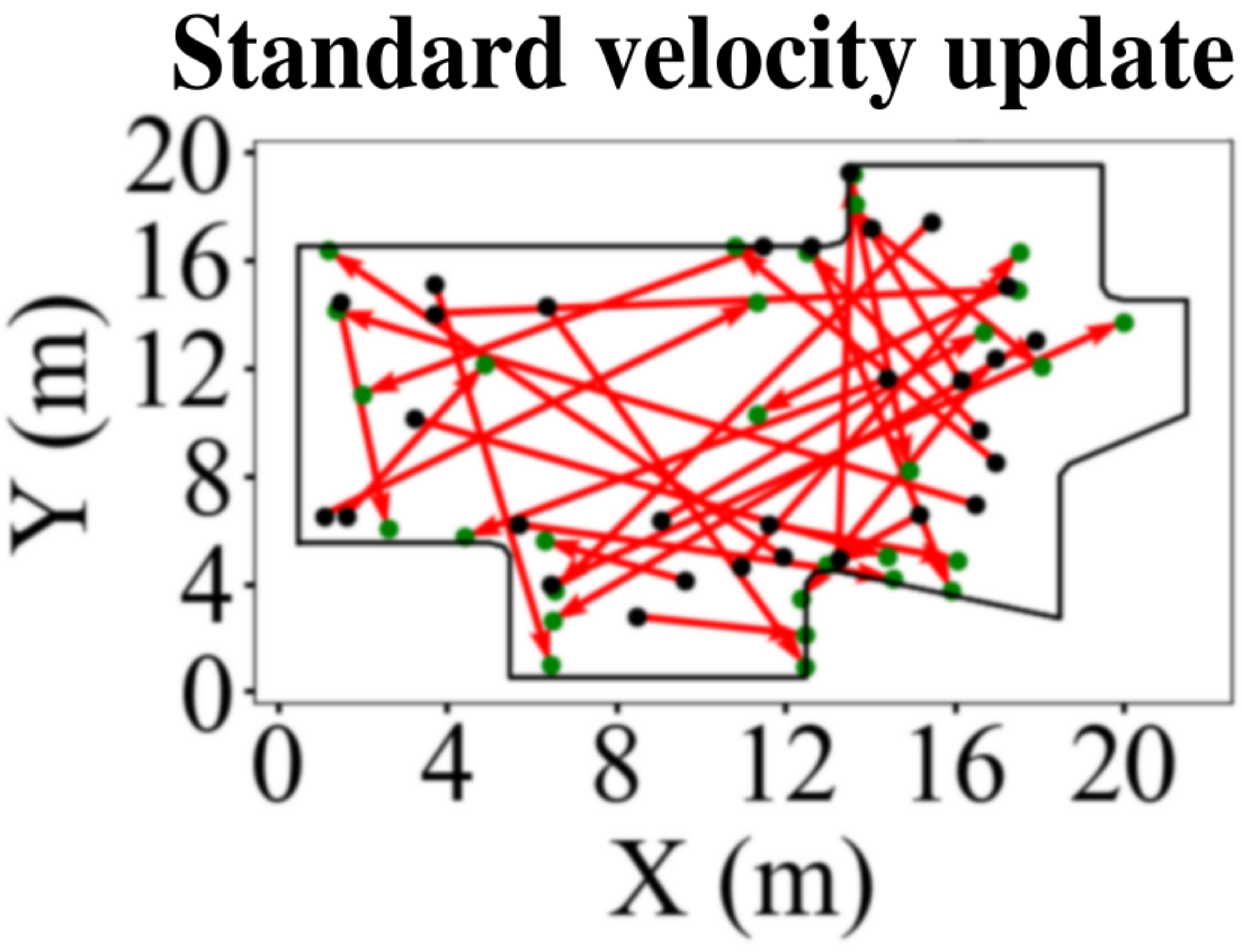}
		\caption{Large vectors and messy velocity updates. }
		\label{fig:psov}
	\end{subfigure}
	\hfill
	\begin{subfigure}[htp]{.25\textwidth}
		\includegraphics[width=\textwidth]{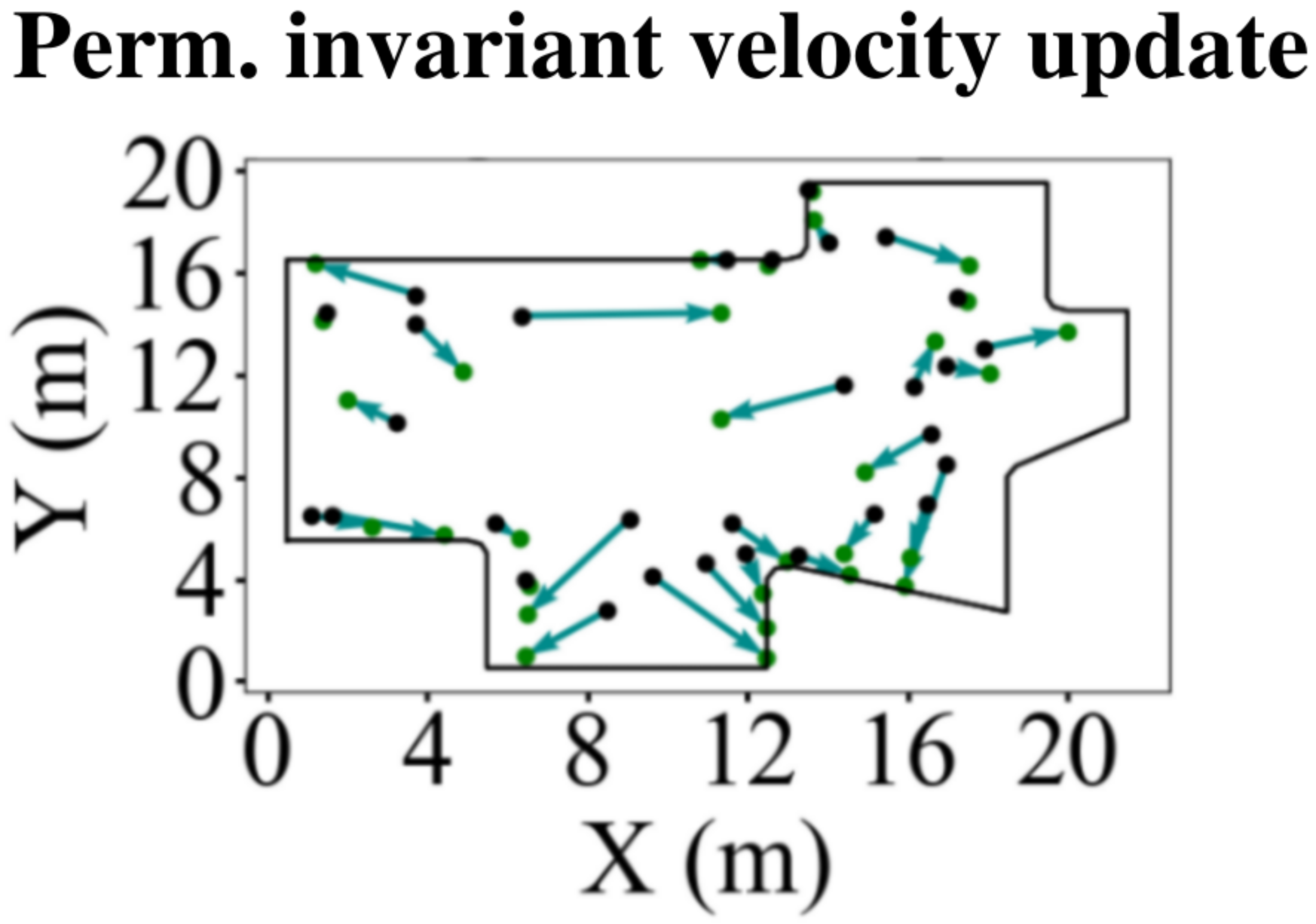}
		\caption{Assignment leads to meaningful velocity updates.}
		\label{fig:permpsov}
	\end{subfigure}
	\caption{Vectors pointing from the particle LRP placement (black dots) to a leader (\emph{pbest} or \emph{gbest}) LRP placement (green dots) that are used to update the particles' velocity vector. The permutation invariant update that assigns particle LRPs to close leader LRPs makes smooth optimization possible.}
	\label{fig:vel_update}
\end{figure}

Another crucial improvement of the standard PSO is the velocity update. As evident from Eq.~\ref{eq:velocity}, we calculate vectors that point from the 2D coordinates of LRPs in a particle to the 2D coordinates of LRPs with the same LRP index in the \emph{pbest} and \emph{gbest} leader solutions. However, LRPs of the same index in those solutions can be anywhere in the room. This may cause large velocity vectors and a messy optimization result with many boundary condition violations.
A better way to perform the velocity update is to compare the LRPs not by index, but to move the LRPs of a particle towards close LRPs in the leader solutions. This can be achieved by permuting the LRPs in a leader with a permutation $\pi_1$ such that the summed distance between LRPs in the particle and the leader is minimized. We solve this assignment problem using the Hungarian method~\cite{hungarian, perm_ga}. Fig.~\ref{fig:vel_update} shows the velocity vectors in a normal PSO velocity update and the superior permutation invariant version. This permutation invariant update has been first proposed for a coverage problem and the~\gls{GA} used as solver in~\cite{perm_ga}. Further, this permutation depends on the positioning mode that we want to optimize. For~\gls{MLAT}, arbitrary reordering of the LRPs in a placement gives equivalent solutions. Therefore, $\pi_1$ is computed w.r.t. all LRPs in the particle and leader. For fingerprinting with only one LRP type, the same applies. If we use more than one LRP type, only reordering of LRPs of the same type leads to equivalent solutions. Exchanging the indices of LRPs of different types results in non-identical and in most cases worse fingerprints. 

Note that we have introduced additional improvements on PSO. They are not described in the paper due to limited space.

\section{Positioning and tracking with AMCL}
Besides the LRP placement for large rooms with arbitrary geometry in section~\ref{sec:pso}, we adopt the positioning and tracking method based
on AMCL from our initial study~\cite{pascal}. AMCL is a particle filter that estimates the AMR position with a set of particles\footnote{Don't confuse the particles in the AMCL method for localization with the particles of the PSO algorithm for LRP placement.}. These particles approximate the true distribution of the AMR position, where each particle corresponds to a candidate position. This approximation is required, as the computation of the true distribution requires the evaluation of all possible room positions, which is computationally not tractable for a real-time~\gls{IPS}. To estimate the position, particles are first initialized by randomly placing them in the room. Each particle is then assigned a weight that correlates with the probability that the AMR is in the same position as the particle. This weight is calculated by comparing the information of detected LRPs with the expected LRPs in this particle position. In our case, we use the distances to LRPs and their LRP types for positioning with fingerprinting and the LRP distances and identifiers for multilateration. The AMR position can then be estimated, e.g. by a weighted average of the positions of the particles. After the positioning, particles are resampled, where good particles with large weight are duplicated and particles with small weight are deleted. Following the initialization, the tracking loop starts, consisting of resampling, position prediction, weight update and position estimation. The only addition is the position prediction, where odometry information can be included. We assume that we get odometry data from the motion control of the~\gls{AMRs}. AMLC can handle nonlinear and multimodal tracking problems and is therefore a suitable choice for our problem since we deal with ambiguity in the position estimates. For a detailed description of~\gls{AMCL}, we refer to~\cite{prob_robotics} and our previous work~\cite{pascal}.
\section{Simulation results}
\begin{figure}[h!]
	\centering
		\begin{subfigure}[t]{.15\textwidth}
		\centering	
		\includegraphics[width=\textwidth]{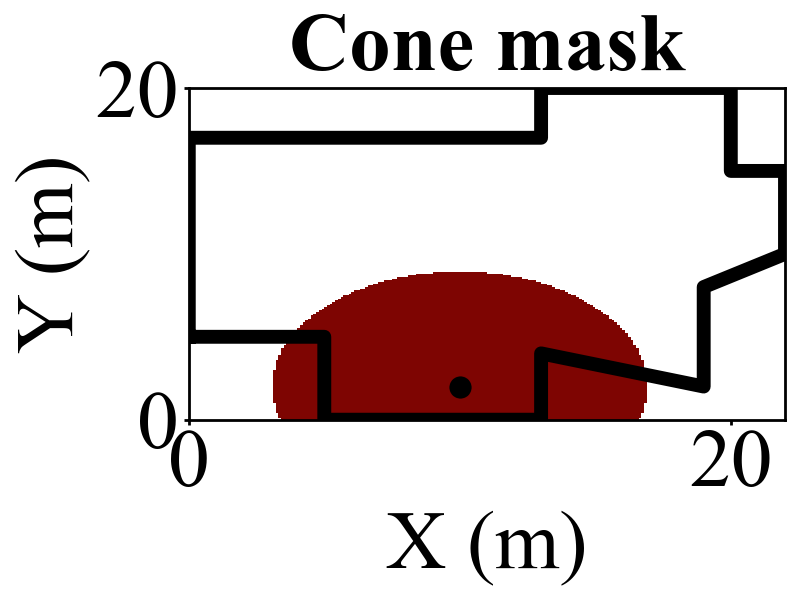}
		\caption{AMR positions (red) where the LRP is in the cone of the radar.}
		\label{fig:cone_mask}
	\end{subfigure}
	\hfill
	\begin{subfigure}[t]{.15\textwidth}
		\centering
		\includegraphics[width=\textwidth]{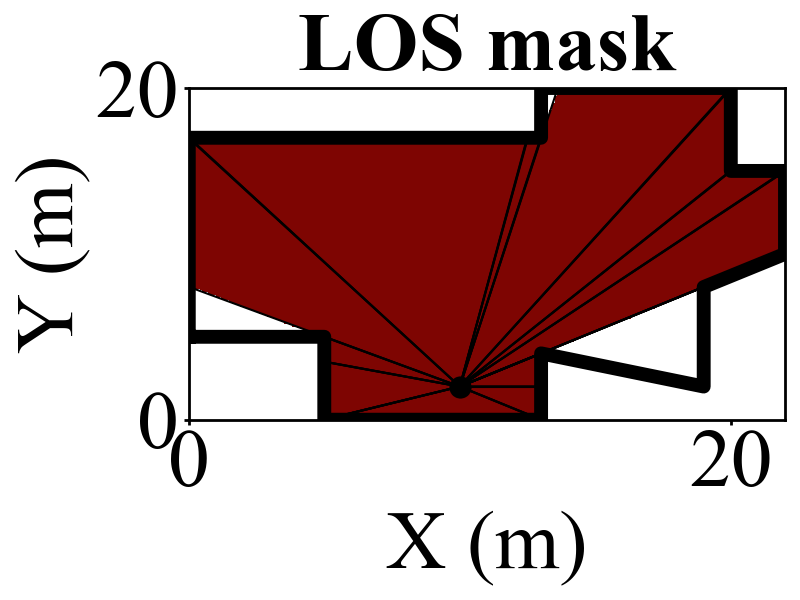}
		\caption{Area with LOS to the LRP computed by raycasting.}
		\label{fig:los_mask}
	\end{subfigure}
	\hfill
	\begin{subfigure}[t]{.15\textwidth}
		\centering
		\includegraphics[width=\textwidth]{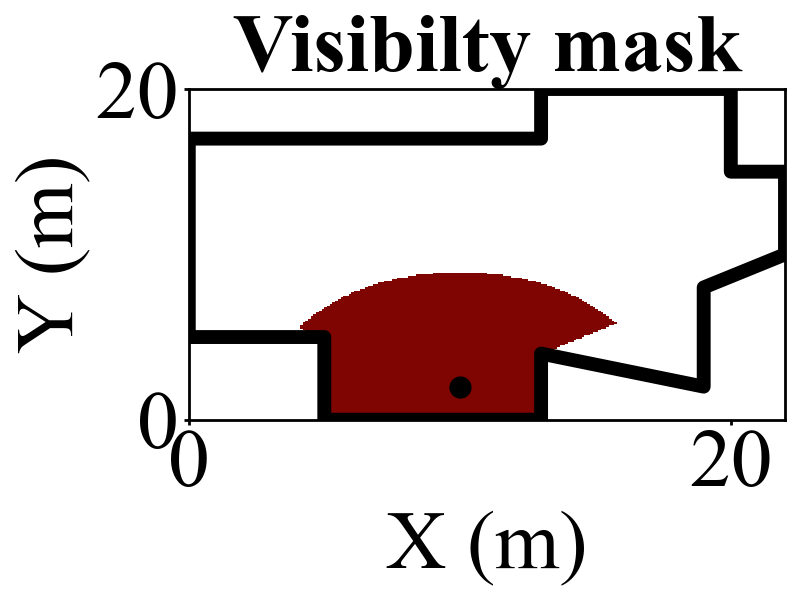}
		\caption{Logical conjunction gives the LRPs' visibility mask.}
		\label{fig:vis_mask}
	\end{subfigure}
	\caption{Binary visibility mask calculation for one~\gls{LRPs} (black dot), which can be detected by the AMR from the red region in c).}
	\label{fig:vis}
\end{figure}

In our simulation, only LRPs that are visible to the radar are available for positioning. The radiation pattern of a radar antenna and the monostatic~\gls{RCS} of realistic reflectors will limit the angle under which reflectors can be detected. We simulate this by spanning a cone from the radar position to the ceiling and check if a LRP lies inside, as shown in Fig.~\ref{fig:cone} and Fig.~\ref{fig:cone_mask}. Further, visibility is only given in~\gls{LOS} conditions. For this, we calculate a visibility polygon for each LRP by casting rays from the LRP to all room vertices. The first intersection points between the rays and room walls are sorted in counterclockwise direction. They are the vertices of the LRPs' visibility polygon. The enclosed area of this polygon is the~\gls{LOS} mask of the LRP. This process is depicted in Fig.~\ref{fig:los_mask}. Logical conjunction of these binary visibility masks give us the positions from which a LRP is visible, as shown in Fig.~\ref{fig:vis_mask}.

In this limited LRP visibility setting, we perform one exemplary optimization for LRP placement with one and two LRP types, respectively. Grid elements are of size 10x10 cm. With $500$ particles and $500$ PSO iterations, we have $250.000$ placement evaluations. The radar has a range resolution of $r_{res}=\SI{7.5}{cm}$, i.e. $\SI{2}{GHz}$ bandwidth. We set $M_{max}=32$ and initialize particles with 27-32 LRPs. For fingerprinting, we use the nearest $N=4$ LRPs. Our optimization takes around 5 days with an AMD Ryzen 5 2600X processing unit.

\begin{figure}[h!]
	\centering
	\begin{subfigure}[t]{.23\textwidth}
		\centering
		\includegraphics[width=0.6\linewidth]{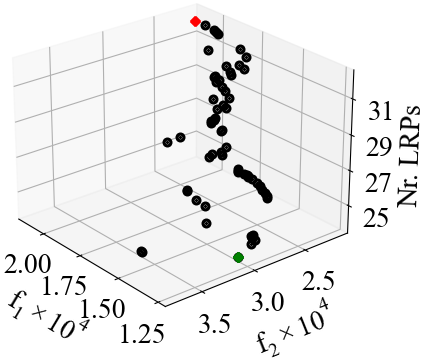}
		\caption{One LRP type: $f_1$ increases and $f_2$ decreases with more LRPs.}
		\label{fig:solutionsfront1}	
	\end{subfigure}
	\hfill
	\begin{subfigure}[t]{.23\textwidth}
		\centering
		\includegraphics[width=0.6\linewidth]{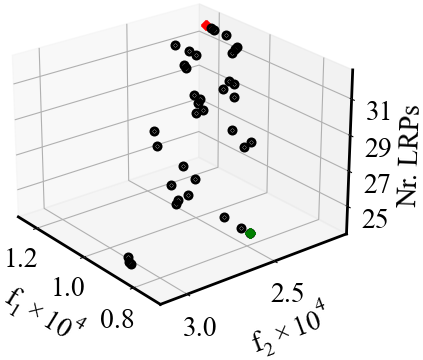}
		\caption{Two LRP types: Smaller $f_1$ but the same objective behavior as with one LRP type.}
		\label{fig:solutionsfront2}	
	\end{subfigure}
	\hfill
	\begin{subfigure}[t]{.23\textwidth}
		\centering
		\includegraphics[width=\textwidth]{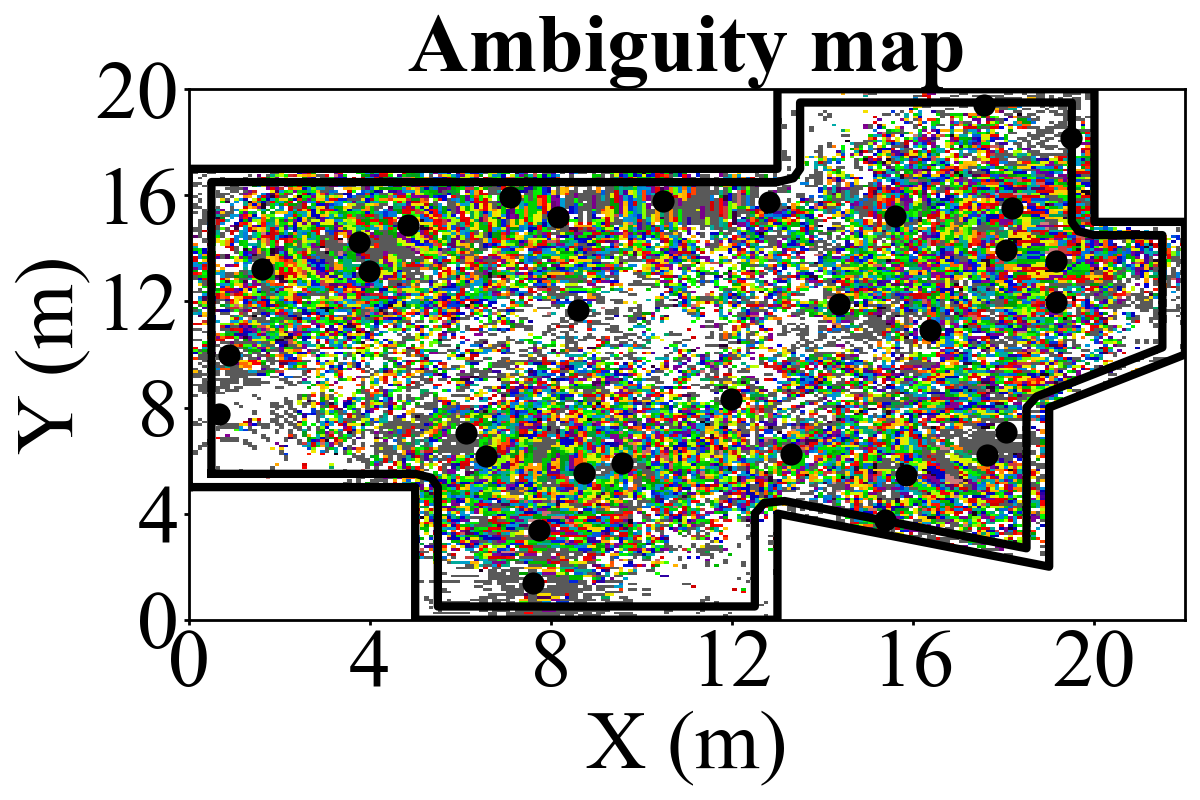}
		\caption{Random LRP placement with one LRP type: $M=32$, $f_1=23358$.}
		\label{fig:amb1_before}
	\end{subfigure}
	\hfill
	\begin{subfigure}[t]{.23\textwidth}
		\centering
		\includegraphics[width=\textwidth]{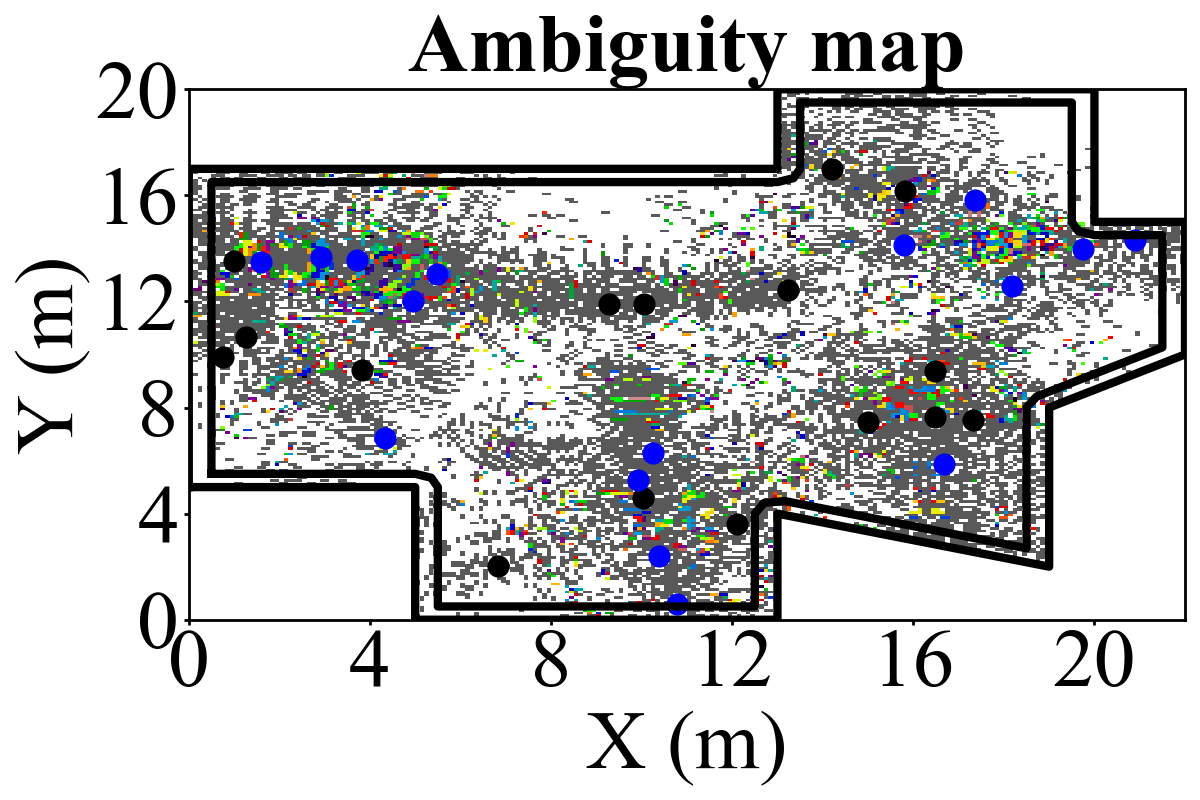}
		\caption{Random LRP placement with two LRP types: $M=32$, $f_1=15994$.}
		\label{fig:amb2_before}
	\end{subfigure}
	\hfill
	\begin{subfigure}[t]{.23\textwidth}
		\centering
		\includegraphics[width=\textwidth]{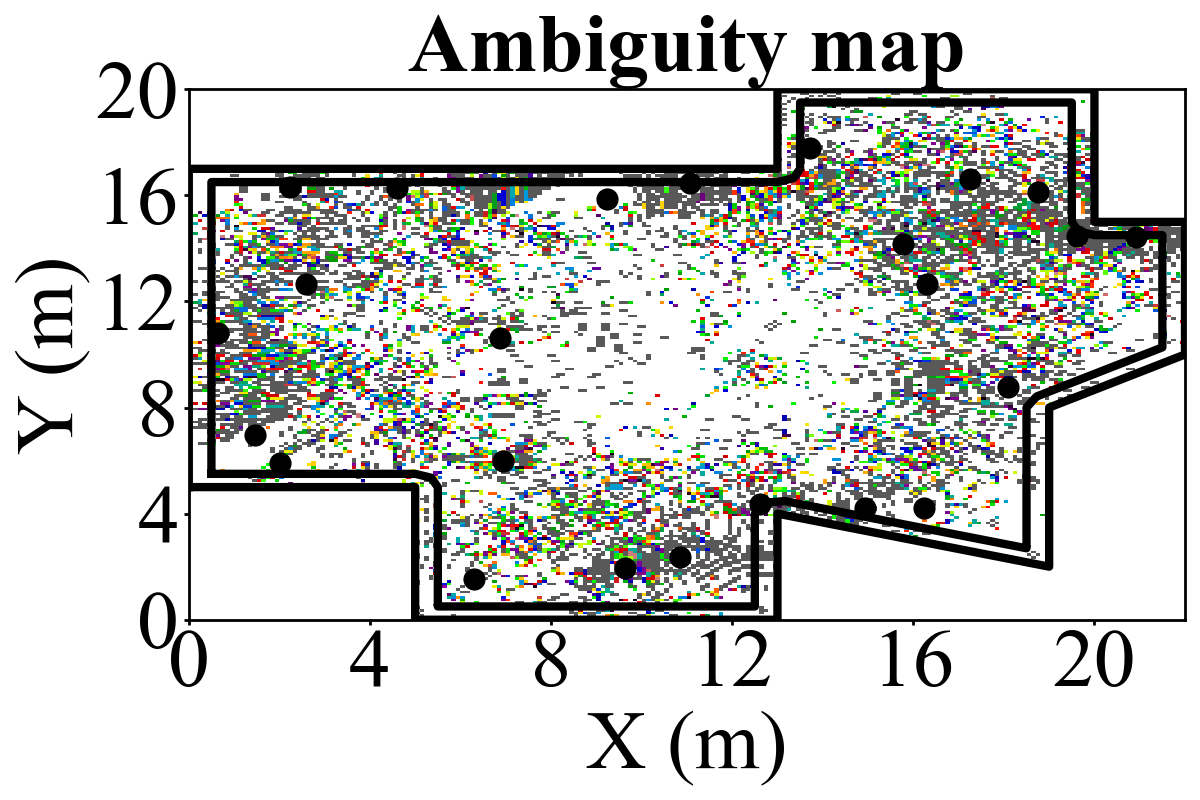}
		\caption{Optimized LRP placement with one LRP type: $M=24$, $f_1=12682$.}
		\label{fig:amb1_after}
	\end{subfigure}
	\hfill
	\begin{subfigure}[t]{.23\textwidth}
		\centering
		\includegraphics[width=\textwidth]{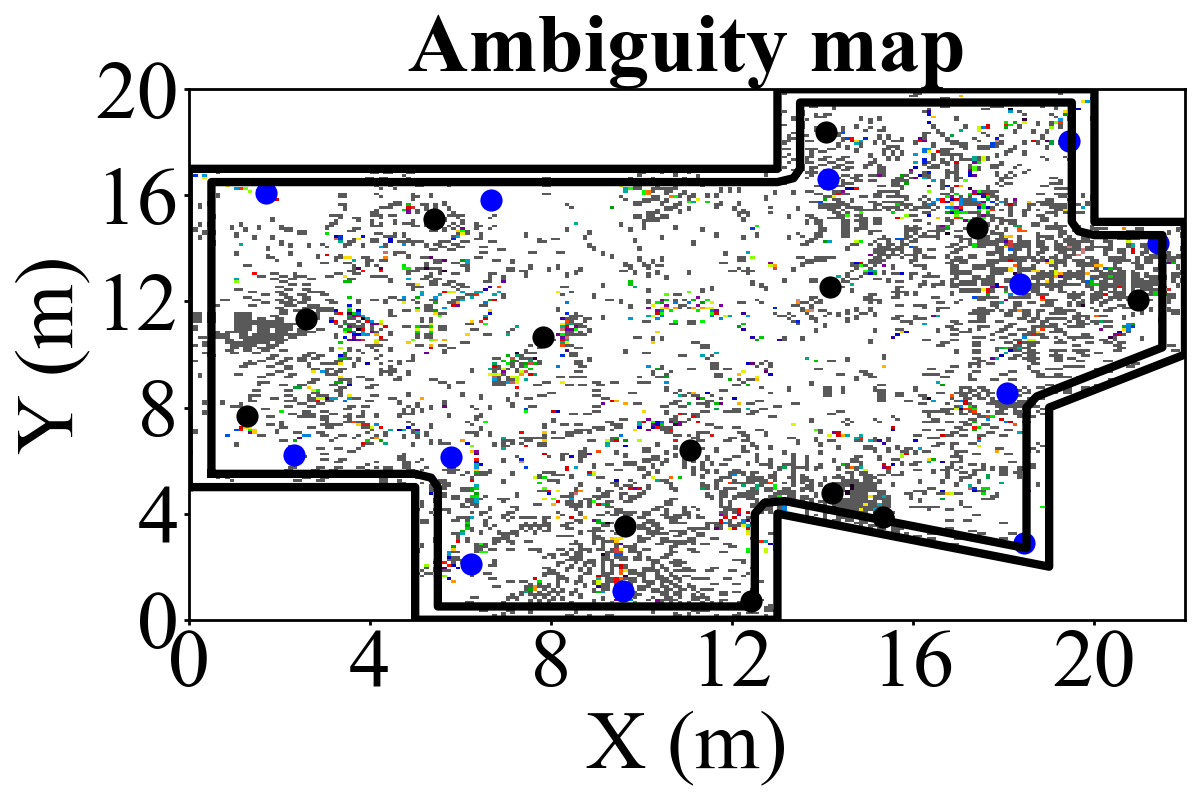}
		\caption{Optimized LRP placement with two LRP types: $M=25$, $f_1=7343$.}
		\label{fig:amb2_after}
	\end{subfigure}
	\hfill
	\begin{subfigure}[t]{.23\textwidth}
		\centering
		\includegraphics[width=\textwidth]{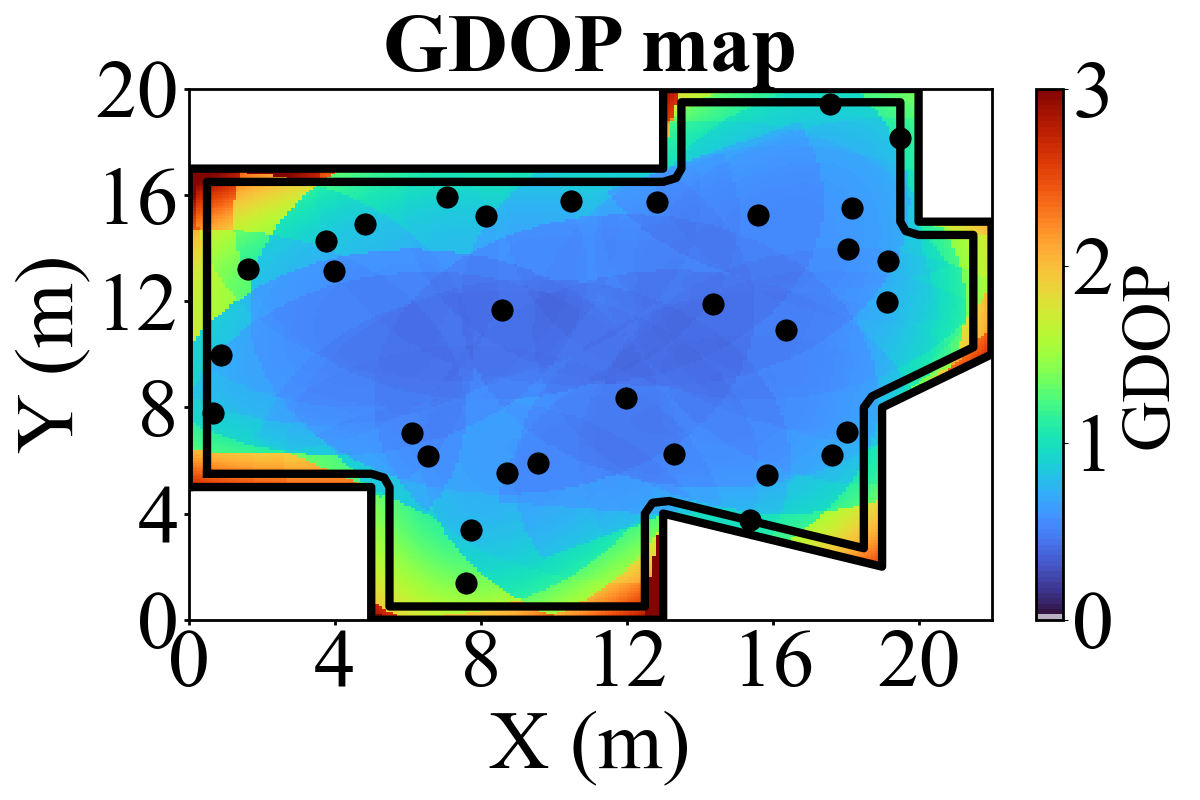}
		\caption{Random LRP placement with one LRP type: $M=32$, $f_2=27622.74$.}
		\label{fig:gdop_before}
	\end{subfigure}
	\hfill
	\begin{subfigure}[t]{.23\textwidth}
		\centering
		\includegraphics[width=\textwidth]{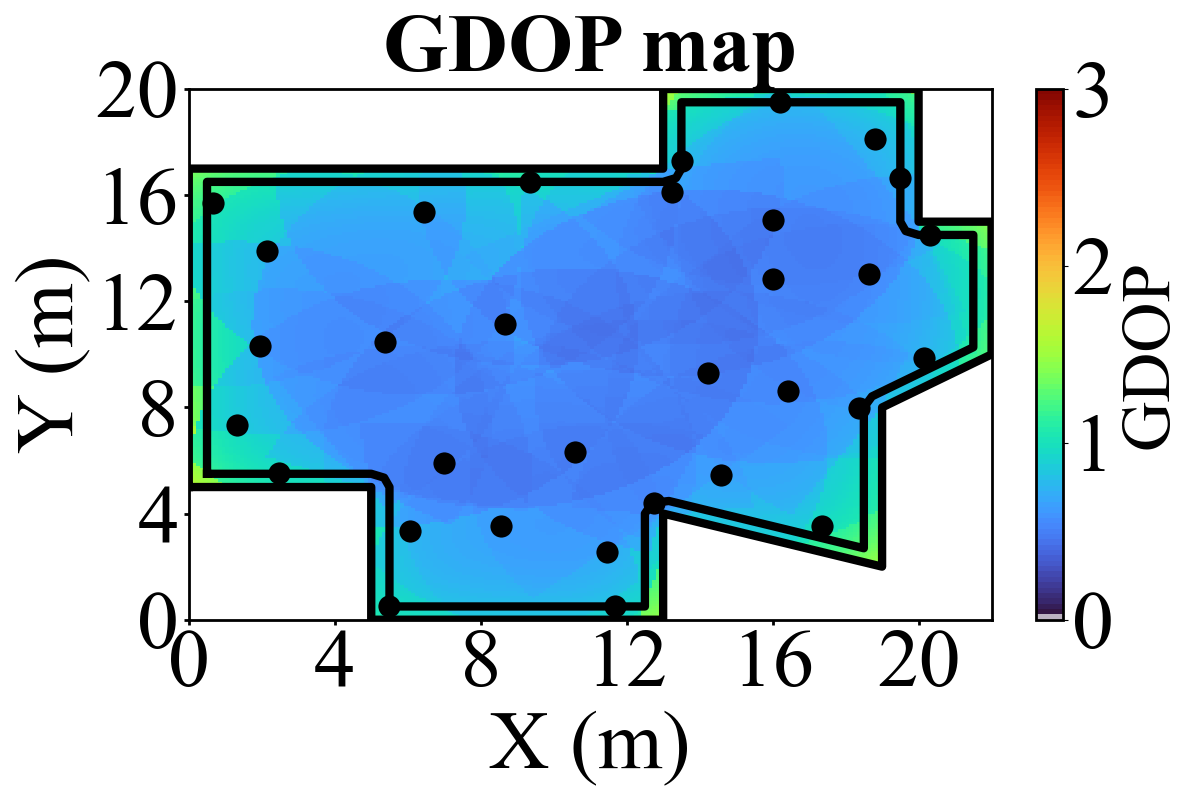}
		\caption{Optimized LRP placement with one LRP type: $M=32$, $f_2=21790.75$.}
		\label{fig:gdop_after}
	\end{subfigure}
	\caption{The sets of optimized solutions for one a) and two b) LRP types show the behavior of $f_1$ and $f_2$ w.r.t. each other and the number of LRPs. For fingerprinting, the best $f_1$ solutions (marked green) are selected and compared with random LRP placements regarding their ambiguous AMR positions (dark gray and colored grid elements) in c) - f). The best MLAT solution $f_2$ for one LRP type (marked red in a)) is used in h) and benchmarked against the GDOP of the random LRP placement for one LRP type (same LRP placement as in c)) in g).}
	\label{fig:results}
\end{figure}

The set of optimized, non-dominated solutions for different numbers of LRPs can be seen in Fig.~\ref{fig:solutionsfront1} and Fig.~\ref{fig:solutionsfront2}. They show the expected decrease in~\gls{GDOP} ($f_2$) with more LRPs, i.e. better~\gls{MLAT} performance. We are also able to reduce the required number of LRPs. This is enabled by our \emph{magnet model} and \emph{gravitation model} and the \emph{downmutation} operation. Brute-force random LRP placement cannot achieve this in reasonable time. Finding a feasible solution in the LRP placement becomes increasingly difficult, the fewer LRPs are used. With 32 LRPs, random LRP placement takes on average already several hundred thousand tries. By using our PSO algorithm, we are able to find feasible solutions with only 24 LRPs.

We further deduct that the fingerprinting approach $(f_1)$ works better with less LRPs. This is expected and caused by the nearest $N$ fingerprint. Using a small total number of LRPs increases the distances between LRPs and therefore reduces the number of ambiguous grid elements. With too few LRPs however, we would have coverage violations and couldn't build a nearest $N$ fingerprint with $N=4$ LRPs. It is also clear, when comparing the $f_1$ scores in Fig.~\ref{fig:solutionsfront1} and Fig.~\ref{fig:solutionsfront2}, that two LRP types lead to a much lower ambiguity $(f_1)$ in fingerprinting based localization. 

To show the effect of optimized LRP placement on $f_1$, we select the best $f_1$ solutions in Fig.~\ref{fig:solutionsfront1} and Fig.~\ref{fig:solutionsfront2} and compare them with random LRP placements with 32 LRPs in Fig.~\ref{fig:amb1_before} to Fig.~\ref{fig:amb2_after}. We only show one random LRP placement per LRP type for comparison. The ambiguity maps in Fig.~\ref{fig:amb1_after} and Fig.~\ref{fig:amb2_after} depict the improvement in $f_1$ after optimization. We differentiate between two kinds of ambiguities, global and local ambiguity. Colored elements in those plots are AMR positions that share the same fingerprint in multiple, unconnected regions in the room. Grid elements with identical fingerprints have the same color. These regions can be located far apart and selecting the wrong one during localization may lead to large positioning errors. However, this problem can be handled by probabilistic, recursive AMR tracking. Dark gray elements form only one connected region of grid elements with a shared fingerprint. This affects the local positioning accuracy. The size of such regions determines the AMR localization accuracy in those positions.

We also compare the~\gls{MLAT} performance of the previous random LRP placement using one LRP type (Fig.~\ref{fig:amb1_before}) and an optimized one by selecting the best $f_2$ solution in Fig.~\ref{fig:solutionsfront1} and plot the GDOP maps of both LRP placements in Fig.~\ref{fig:gdop_before} and Fig.~\ref{fig:gdop_after}. The GDOP, especially in positions near walls, is largely improved, showing the ability of our proposed algorithm to optimize both positioning modes. 

\begin{figure}[h!]
	\centering
	\includegraphics[width=0.99\linewidth]{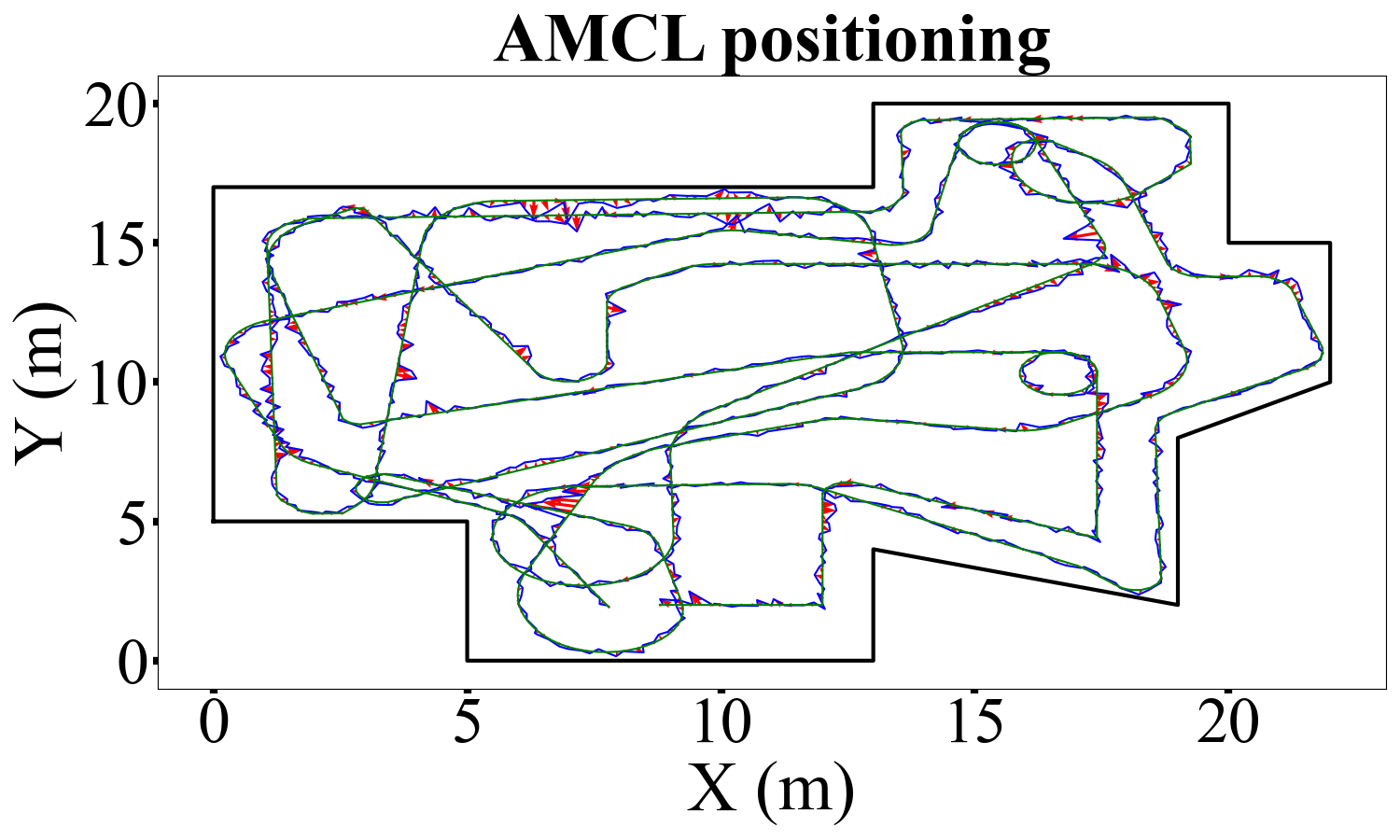}
	\caption{Ground truth AMR test path (green line), prediction by AMCL (blue line) and positioning error vectors (red) using the optimized LRP placement with one LRP type.}
	\label{fig:amcl_path}
\end{figure}

\begin{figure}[h!]
\centering
	\includegraphics[width=0.7\linewidth]{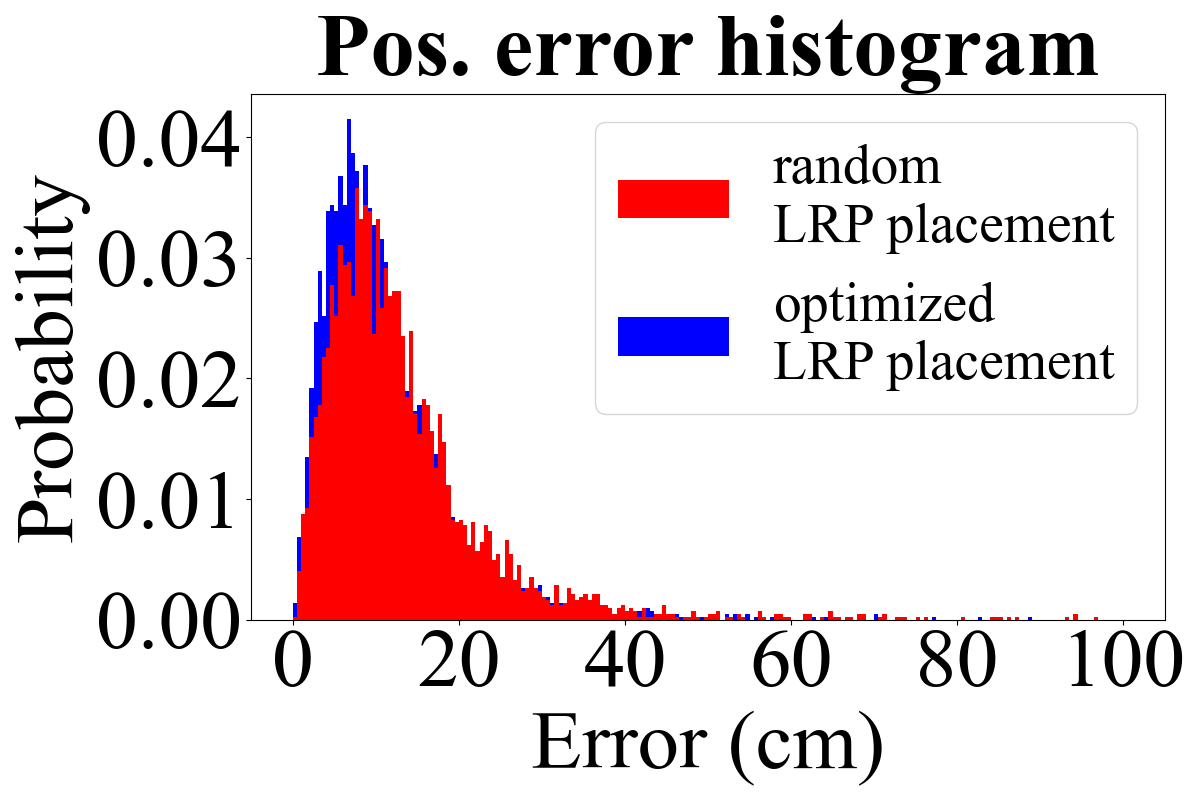}
	\caption{The optimized LRP placement gives notably better positioning performance while requiring only 24 instead of 32 LRPs.}
	\label{fig:amcl_hist}	
\end{figure}

For the AMCL performance evaluation, we create a test AMR motion path and simulate the fingerprinting mode using one LRP type with the LRP placement solutions from Fig.~\ref{fig:amb1_before} and Fig.~\ref{fig:amb1_after}. The AMR path with 1411 total position estimates by AMCL with Fig.~\ref{fig:amb1_after} as LRP placement is shown in Fig.~\ref{fig:amcl_path}. The AMR position is estimated every $\SI{20}{cm}$. To simulate noisy sensors, we add zero mean Gaussian noise with a standard deviation of $\sigma=r_{res}$ to the true LRP distances before rounding them with $r_{res}$ and select the nearest 4 distances as our fingerprint for positioning. We also add zero mean Gaussian noise with a standard deviation of $\sigma_{d}=\SI{2}{cm}$ for the distance and $\sigma_{\theta}=\ang{5}$ for the rotation angle to the true data between AMR movements. The positioning error histograms are found in Fig.~\ref{fig:amcl_hist}. Using the random placement of 32 LRPs, localization with a~\gls{RMSE} of $\SI{20.85}{cm}$ is possible. With the optimized placement of only 24 LRPs, we achieve an~\gls{RMSE} of $\SI{13.54}{cm}$, less than twice of the radar range resolution.
		
\section{Discussion}
This papers presents a conceptual study for indoor localization using a low-complexity single-channel FMCW radar for sensing and simple passive radar reflectors called LRPs. The main focus is the optimal LRP placement in a large room of arbitrary geometry.
Our positioning and tracking results are currently evaluated in an optimistic setting. We use noisy odometry inputs and range measurements, but we assume that the detected LRPs are always correct, i.e. there are no false alarms, miss detections or incorrectly associated LRP types. The performance in reality might be worse and shall be studied in future work. However, as our radar is facing the ceiling, we expect multipath effects to be not as detrimental as in other~\gls{RF} indoor positioning applications. We also do not utilize Doppler information yet and use only the nearest $N=4$ instead of all detectable LRPs for positioning with fingerprinting.
If the differentiability of LRP types by RCS proves to be too difficult, one could use a polarimetric radar and reflectors with different polarimetric behavior. Another option would be a multi-channel system to integrate angular information. To eliminate the global initialization problem, a few active tags could be added to the system as easily detectable and identifiable anchor points.

\section{Conclusion}
We have extended our previous work~\cite{pascal} on a novel indoor positioning concept based on detections of local, passive radar reflectors with a simple single-channel~\gls{FMCW} radar. While only requiring low-complexity components, we can achieve high accuracy by using chirps of large bandwidth and recursive AMR tracking.

We introduced two positioning modes for our system, fingerprinting and~\gls{MLAT}. We introduced various improvements to the basic~\gls{PSO} algorithm for optimum LRP placement: the ability to optimize both positioning modes with different numbers of~\glspl{LRPs} in a single optimizer, the handling of boundary condition violations with our proposed physical models, and the permutation invariant velocity update etc.

The optimization not only improves set of objectives through better LRP placements, but also allows to choose the preferred trade-off between the number of LRPs and both positioning modes. We could further demonstrate the advantage of using optimized LRP placements by tracking a simulated AMR in the room with our AMCL solution. 

We plan to integrate more realistic detection scenarios in future work. This will be done by either raytracing to generate synthetic indoor channels as in our previous work~\cite{pascal} or with a real experimental system. Also, radar false alarm and miss will be considered. In addition, the~\gls{SLAM} like~\gls{MLAT} mode will be implemented and compared to fingerprinting.
\bibliographystyle{ieeetr}
\bibliography{ref}

@inproceedings{omopso,
	author = {Sierra, Margarita and Coello, Carlos},
	year = {2005},
	month = {},
	pages = {},
	title = {Improving {PSO}-Based Multi-objective Optimization Using Crowding, Mutation and  $\epsilon$-Dominance},
	volume = {3410},
	isbn = {978-3-540-24983-2},
	journal = {Lecture Notes in Computer Science},
	doi = {10.1007/978-3-540-31880-4_35}
}

@INPROCEEDINGS{pascal,
	
	author={Pascal Schlachter and others},
	
	
	title={Indoor Positioning based on Active Radar Sensing and Passive Reflectors: Concepts and Initial Results}, 
	
	year={2023},
	
	volume={},
	
	number={},
	
	pages={},
	}

@book{prob_robotics,
	address = {Cambridge, Mass.},
	author = {Thrun, Sebastian and Burgard, Wolfram and Fox, Dieter},
	description = {Probabilistic Robotics Intelligent Robotics and Autonomous Agents: Amazon.de: Sebastian Thrun, Wolfram Burgard, Dieter Fox: Bücher},
	isbn = {0262201623 9780262201629},
	publisher = {MIT Press},
	refid = {58451645},
	title = {Probabilistic robotics},
	url = {http://www.amazon.de/gp/product/0262201623/102-8479661-9831324?v=glance&n=283155&n=507846&s=books&v=glance},
	year = 2005
}

@Article{uwb,
	AUTHOR = {Alarifi, Abdulrahman and others},
	TITLE = {Ultra Wideband Indoor Positioning Technologies: Analysis and Recent Advances},
	JOURNAL = {Sensors},
	VOLUME = {16},
	YEAR = {2016},
	NUMBER = {5},
	ARTICLE-NUMBER = {707},
	URL = {https://www.mdpi.com/1424-8220/16/5/707},
	PubMedID = {27196906},
	ISSN = {1424-8220},
	DOI = {10.3390/s16050707}
}

@INPROCEEDINGS{fmcwindoor,

author={Marcaccioli, Luca and others},

booktitle={IEEE Intelligent Vehicles Symposium (IV)}, 

title={An accurate indoor ranging system based on {FMCW} radar}, 

year={2011},

volume={},

number={},

pages={981-986},

doi={10.1109/IVS.2011.5940477}}

@inproceedings{millimetro,
	author = {Soltanaghaei, Elahe and others},
	title = {Millimetro: MmWave Retro-Reflective Tags for Accurate, Long Range Localization},
	year = {2021},
	isbn = {9781450383424},
	publisher = {},
	address = {},
	url = {https://doi.org/10.1145/3447993.3448627},
	doi = {10.1145/3447993.3448627},
	booktitle = {Proceedings of the 27th Annual International Conference on Mobile Computing and Networking},
	pages = {},
	numpages = {14},
	location = {New Orleans, Louisiana},
	series = {}
}

@inproceedings{vforce,
	author = {Zou, Yi and Chakrabarty, Krishnendu},
	year = {2003},
	month = {},
	pages = {1293 - 1303},
	title = {Sensor deployment and target localization based on virtual forces},
	volume = {2},
	isbn = {0-7803-7752-4},
	journal = {Proceedings of the IEEE INFOCOM},
	doi = {10.1109/INFCOM.2003.1208965}
}

@ARTICLE{perm_ga,
	
	author={Yoon, Yourim and Kim, Yong-Hyuk},
	
	journal={IEEE Transactions on Cybernetics}, 
	
	title={An Efficient Genetic Algorithm for Maximum Coverage Deployment in Wireless Sensor Networks}, 
	
	year={2013},
	
	volume={43},
	
	number={5},
	
	pages={1473-1483},
	
	doi={10.1109/TCYB.2013.2250955}}

@article{hungarian,
	author = {Kuhn, H. W.},
	title = {The Hungarian method for the assignment problem},
	journal = {Naval Research Logistics Quarterly},
	volume = {2},
	number = {1-2},
	pages = {83-97},
	doi = {https://doi.org/10.1002/nav.3800020109},
	url = {https://onlinelibrary.wiley.com/doi/abs/10.1002/nav.3800020109},
	eprint = {https://onlinelibrary.wiley.com/doi/pdf/10.1002/nav.3800020109},
	year = {1955}
}

@INPROCEEDINGS{pso,
	
	author={Kennedy, J. and Eberhart, R.},
	
	booktitle={Proceedings of International Conference on Neural Networks}, 
	
	title={Particle swarm optimization}, 
	
	year={1995},
	
	volume={4},
	
	number={},
	
	pages={1942-1948},
	
	doi={10.1109/ICNN.1995.488968}}

@article{nsga2,

author={Deb, K. and Pratap, A. and Agarwal, S. and Meyarivan, T.},

journal={IEEE Transactions on Evolutionary Computation}, 

title={A fast and elitist multiobjective genetic algorithm: {NSGA-II}}, 

year={2002},

volume={6},

number={2},

pages={182-197},

doi={10.1109/4235.996017}}

@article{vndpso,
	author = {Kadlec, Petr and \v{S}Šed\v{e}ěnka, Vladimír},
	year = {2017},
	month = {},
	pages = {1-18},
	title = {Particle swarm optimization for problems with variable number of dimensions},
	volume = {50},
	journal = {Engineering Optimization},
	doi = {10.1080/0305215X.2017.1316845}
}

@article{gdop3drange,
	author = {Li, Binghao and others},
	year = {2011},
	month = {},
	pages = {343-349},
	title = {{3D DOPs} for Positioning Applications Using Range Measurements},
	volume = {3},
	journal = {Wireless Sensor Network},
	doi = {10.4236/wsn.2011.310037}
}

@INPROCEEDINGS{nsga2vsomopso,
	
	author={Godínez, Adriana Cortés and Espinosa, Luis Ernesto Mancilla and Montes, Efren Mezura},
	
	booktitle={IEEE Electronics, Robotics and Automotive Mechanics Conference}, 
	
	title={An Experimental Comparison of Multiobjective Algorithms: {NSGA-II} and {OMOPSO}}, 
	
	year={2010},
	
	volume={},
	
	number={},
	
	pages={28-33},
	
	doi={10.1109/CERMA.2010.13}}
\end{document}